\documentclass{article} 
\usepackage{iclr2020_conference,times}

\usepackage{amsmath,amsfonts,bm}

\def\eqref#1{equation~\ref{#1}}

\def\1{\bm{1}}

\DeclareMathAlphabet{\mathsfit}{\encodingdefault}{\sfdefault}{m}{sl}
\SetMathAlphabet{\mathsfit}{bold}{\encodingdefault}{\sfdefault}{bx}{n}

\usepackage{hyperref}
\usepackage{amsmath}
\usepackage{bm}
\usepackage[noabbrev]{cleveref}
\usepackage{xcolor}
\usepackage{textcomp}
\usepackage{graphicx}
\usepackage{tabularx}
\usepackage[font=small]{caption}
\usepackage{subfig}
\usepackage{csquotes}
\usepackage{amssymb}
\usepackage{algorithm}
\usepackage[noend]{algpseudocode}
\makeatletter
\def\BState{\State\hskip-\ALG@thistlm}
\makeatother
\usepackage{rotating}
\usepackage{verbatim}
\usepackage{geometry}
\usepackage{pdflscape}
\usepackage{multicol}
\usepackage{etoolbox}
\usepackage{url}

\title{Graceful Degradation and Related Fields}

\author{Jack Dymond \\
School of Electronics and Computer Science\\
University of Southampton\\
\texttt{j.dymond@soton.ac.uk} \\
}

\iclrfinalcopy 
\begin{document}

\maketitle
\begin{abstract}

When machine learning models encounter data which is out of the distribution on which they were trained they have a tendency to behave poorly, most prominently over-confidence in erroneous predictions. Such behaviours will have disastrous effects on real-world machine learning systems. In this field graceful degradation refers to the optimisation of model performance as it encounters this out-of-distribution data. This work presents a definition and discussion of graceful degradation and where it can be applied in deployed visual systems. Following this a survey of relevant areas is undertaken, novelly splitting the graceful degradation problem into active and passive approaches. In passive approaches, graceful degradation is handled and achieved by the model in a self-contained manner, in active approaches the model is updated upon encountering epistemic uncertainties. This work communicates the importance of the problem and aims to prompt the development of machine learning strategies that are aware of graceful degradation.

\end{abstract}
\newpage
\tableofcontents
\newpage

\section{Introduction}

Over the last decade the amount of data that is available in our society has increased massively, from football match statistics to personal information, data has become another currency with which companies compete with one another. With this data often being unstructured, the necessity for analysts and data scientists has increased across every industry in the world. As such, the tools and algorithms that they use have seen significant surges in development, both in the academic and commercial environment. The field of machine learning has been progressing at an unprecedented rate, this progress being led by research staff at universities across the world. Heavy investment from tech giants such as Google and Microsoft and newcomers such as DeepMind and OpenAI has prompted collaboration with universities, allowing greater acceleration in progress. Hardware manufacturers such as ARM and Nvidia have shifted towards giving AI a central role in their business models, hence, both ends of the technology industry are contributing resources to the field that help facilitate this surge in progress. 

The data passed to these systems has increased in volume and complexity, making the manual extraction of features unfeasible. Consequently, the less complicated algorithms that once dominated the field have seen less development. In their place deep neural networks have seen a significant rise and have sparked what is often referred to as a ‘Golden Age’ of machine learning. Deep learning is effective because of its unparalleled ability to find meaningful features and representations in data and its ability to generalise these across large amounts of data. 

This has prompted much interest from many industries around the world, from the defence to the pharmaceutical sector, companies are forming the same academic alliances as tech companies to help improve the products and services they can offer. Indeed this has sparked many developments that will be influential for years to come, for example DeepMind's AlphaFold2 (\cite{Jumper2020}), a deep learning system that has solved the 50-year-old protein folding problem in biochemistry first presented by \cite{levinthal1969fold}. 

But these systems are not without flaws. For example DeepMind's other groundbreaking system, AlphaGo, predicted with high certainty that it would win the one match it lost against the world champion Go player. That is until a pivotal move changed the game (\cite{silver2017mastering}). 

This is not something limited to these state-of-the-art systems, but is a problem of many deep learning systems. In vision systems, which will serve as the focus of this survey, these effects might occur when encountering data outside of the training distribution. These will cause uncalibrated models to output over-confident, yet erroneous predictions (\cite{Li_2020_CVPR}). When deployed in safety-critical environments such mistakes will have catastrophic consequences, thus in many environments it is critical that the machine learning systems be implemented in a safe and responsible manner. To utilise deep learning in sectors that have real-world consequences it is important to ensure these shortcomings are managed responsibly. To that end, such techniques should be integrated into systems with the concept of graceful degradation built into them.

\subsection{Defining Graceful Degradation}

Outside of the machine learning environment graceful degradation refers to the notion of making a system resilient to disruption in its architecture (\cite{herlihy1991specifying}). That is, when parts of the system are inoperable, the system’s performance will degrade in the most efficient way possible. 

In machine learning, achieving graceful degradation can be interpreted as optimising the drop in performance of a model as it processes data that is further from the distribution on which it was trained. These are commonly known as out-of-distribution errors(\cite{hendrycks2016baseline}). Graceful degradation is the notion of minimising their effect. 

The problem with out-of-distribution errors when applying machine learning is not the errors themselves, it is the manner in which such errors occur. State-of-the-art machine learning techniques can achieve beyond human accuracy on some tasks as demonstrated by the the pursuit of high accuracies in the ImageNet challenge, currently being led by \cite{dosovitskiy2020image}. However, when such models are presented images outside of their training distributions, they are often susceptible to making over-confident erroneous predictions (\cite{Li_2020_CVPR}).

There are a variety of ways one can correct for such effects. Fundamentally, the model should understand the class distribution on which it is trained and know when it encounters something outside of it. That is, the model should understand when it is uncertain. 

Then the way in which the model deals with uncertainty defines the approaches that can be taken in achieving graceful degradation. For example, the operator may wish for the model to appropriately signify when it is low on confidence and do this consistently when encountering out-of-distribution data. Alternatively, the operator may instead wish for the model to function on out-of-distribution data and have it trained to recognise objects without having prior examples of them, a method known as zero-shot learning. There could also be a compromise between these approaches whereby the model does not offer a fine-grained classification, but instead offers a coarser-level classification, recognising the hierarchical nature of its inputs. Alternatively, it may be more desirable for the model retrain on the unseen distribution. This might be in an individual data point manner by identifying areas in the distribution that lack the desired sample density, it might be shifting to an entirely new domain of data, or it may be training on an entirely different problem. 

To give some of these methodologies context, a number of vignettes are detailed below that can help provide some clarity. 

\paragraph{Disaster Response Drone}

A disaster response drone might be trained to recognise various scenes in natural disaster scenarios as a way of quickly surveying an area involved in such a disaster. For example, the surrounding coastline in the aftermath of a tsunami may need to be surveyed for notable features such as flooded areas, collapsed buildings, and also unaffected areas to inform sensible allocation of resources. The visual system behind such a drone could be trained in a number of ways. It could be desirable to inform the operator when it is processing images outside of its expertise. Here, the model could express its uncertainty in such scenarios and recommend manual cataloguing of the area. Alternatively, it may be given explicit examples of types of vehicles, but only semantic information of other types and is expected to classify these should it come across them, highlighting the classification as zero-shot. Finally, it might instead be required to give a partial classification, such as the category of an object e.g., it might recognise an area of land is flooded, but may not be able to infer the type of land that is underneath. (Some examples of vision models in drone systems are given in: \cite{8326187}, \cite{Gemert2015} ,\cite{radovic2017object})

\paragraph{Smart Cities}

Cities process large amounts of data every day and as smart city systems become more prevalent, manually processing the data will become intractable. Hence, it may be beneficial to automatically process some of the data seen by the system. The system may encounter information that is not within its training distribution, that it could not process confidently. This could be handled in a variety of ways. The information could be communicated to and processed by an operator, which would allow the model to update its parameters in this area of the distribution. The model could also be having difficulty due to seasonal changes in the information, such as increased traffic flow around public holidays and can be retrained on examples of this. Finally, the model could be retrained to recognise situations with different volumes of pedestrians and the changes that come with them, such as areas which have an increased chance hooliganism at times surrounding football matches. Ideally, a single system would be able to identify all of the aforementioned scenarios, but this is not always practical due to computational constraints in a lightweight system. Hence, there may be a need to retrain the model in an intelligent manner. (Some examples of artificial intelligence models in a similar domain are given in: \cite{ai2020extreme}, \cite{kihara2019designing})

The applications of graceful degradation can be split into two sections, one where the system deals with out-of-distribution data in a self-contained manner, and one where the system can take additional inputs to improve its performance in the observed distribution. These two approaches can be split into passive and active approaches.

\subsection{Passive vs Active}

When dealing with out-of-distribution data a gracefully degrading system should be able to operate in a desirable manner. Broadly speaking, this can come in one of two ways: One whereby the model identifies it as out-of-distribution and gives an accordingly conditioned output, then another whereby the model requires additional information in order to continue operating effectively. These are passive approaches and active approaches, respectively.

In passive approaches to graceful degradation, the model is a self-contained system that can provide a solution to the graceful degradation problem without requiring further input. An example of a passive approach might include simply providing good uncertainty estimates in the drone vignette, whereby the system informs the operator when it is uncertain about what it is seeing. Such approaches will be considered in \cref{epistemic-uncertainties}. Another example might be where the model can attempt to classify out-of-distribution data without ever being shown examples of it, such zero-shot classification is discussed in \cref{zero-shot}. Finally, there might be a middle ground between these two, whereby the model cannot classify an item, but it can go some way to indicate what it is viewing, taking into account the class hierarchy and giving a coarser-grained prediction. This is discussed in \cref{hierarchical-classification}.

In contrast, rather than remain static, active approaches seek to improve their understanding of the task and its data once encountering uncertainties. This will be done by updating or retraining the model. The degree to which this update changes can separate the active approaches. The model might lack understanding in a particular area of the data distribution and require annotation from an operator to inform a parameter update. In \cref{active-continual} these small changes will be discussed, in relation to active and continual learning paradigms in the field. The model might be encountering too much out-of-distribution data for small updates to help, hence it may need to be completely retrained for a new distribution/domain. This problem will be discussed in the context of domain adaptation in \cref{domain-adaptation}. Finally, the system may now be encountering completely new data thus changing its task. Here, models would be required to quickly learn new tasks and this could be solved by meta-learning. These approaches will be discussed in \cref{meta-learning}.

A core idea in all of these scenarios is the model understanding when it is uncertain and out of its depth. Hence, it will be important to understand the various ways of defining uncertainty. This will be covered briefly in the next section. 


\subsection{Understanding Uncertainty} \label{uncertainties}

In order to investigate uncertainties in classification, it is important to first understand the various interpretations of uncertainty in the machine learning field and how they can be quantified, starting with the latter. 

To understand a model's uncertainty, it needs to be quantified. There are variety of ways to do this and in the deep learning field usually the output of the model is used as a means of obtaining the uncertainty. This output is most commonly in the form of a softmax distribution representing a probability over the class space, a technique popularised by \cite{duan2003multi}. This is presented mathematically in \cref{softmax}, where $p(y_i)$ is the probability of class $i$ and $N$ is the number of classes. Here, $y$ refers to the output of a neural network's last layer, representing prediction values for each , $i$, class in a classification task. The softmax function converts these raw numbers into probabilities.

This probability distribution is often passed into the cross entropy formula, which was first presented by \cite{10.2307/2984087}. This is shown in \cref{cross-entropy}, where, $q$ is the estimation from the machine learning algorithm, often using a soft-max, and $p$ is the statistical probability accounting for the distribution of classes in the data. These are often used as confidence scores, something that can be interpreted as the inverse of uncertainty. 

\begin{tabularx}{\textwidth}{XX}
\begin{equation}\label{softmax}
p(y_{i}) = \frac{e^{y_i}}{ \sum\limits_{j=1}^{N}  e^{y_j}}
\end{equation}
    &
\begin{equation} \label{cross-entropy}
I(p,q) = -\sum_{i=1}^{N} p(y_i) \log q(y_i)
\end{equation}
\end{tabularx}

There also exist a variety of evaluation measures in the field that are related to confidence and will use confidence scores in their calculation. For example, the E99 error is the error rate in classifications where the classifier records 99\% certainty in its prediction. This is a quantity discussed in a recent paper by \cite{Li_2020_CVPR}, which analyses over-confidence in erroneous classifications on out-of-distribution data. Expected calibration error (ECE), which was first described as such by \cite{naeini2015obtaining}, measures whether or not the classifier ``knows what it knows", as it is designed to measure the alignment between the accuracy and the confidence by summating the difference between them over given intervals. There also exists the Brier score, first developed for weather prediction by \cite{brier1950verification}. In this score the difference is taken between the confidence in the correct the label and the number 1. Hence, this is optimised when the confidence in the correct label is maximised. Such measurements are useful in the field of graceful degradation, as they emphasise where confidence is misplaced. Hence, a variety of confidence calculation methods are compatible with gracefully degrading systems.

Uncertainty does not only exist within the trained distribution. In the graceful degradation problem it is the uncertainty on data outside of the trained distribution that is more important. These types of uncertainty are known as aleatoric and epistemic uncertainty (\cite{hullermeier2020aleatoric}). In machine learning, aleatoric uncertainty is the uncertainty arising due to the natural stochasticity in the data, it can be likened to the ability for the model to interpolate effectively. On the other hand, epistemic uncertainty arises due the training data insufficiently representing the data being observed by the model. That is, the model is encountering data it is not familiar with. This might be due to insufficient training data in the area, or it might be due to the data being entirely out-of-distribution, and the model has failed to extrapolate effectively. Hence, epistemic uncertainty is the most relevant error in the graceful degradation problem as it concerns error due to out-of-distribution samples. 

\begin{figure}[h!]
\centering
\includegraphics[width=0.6\textwidth]{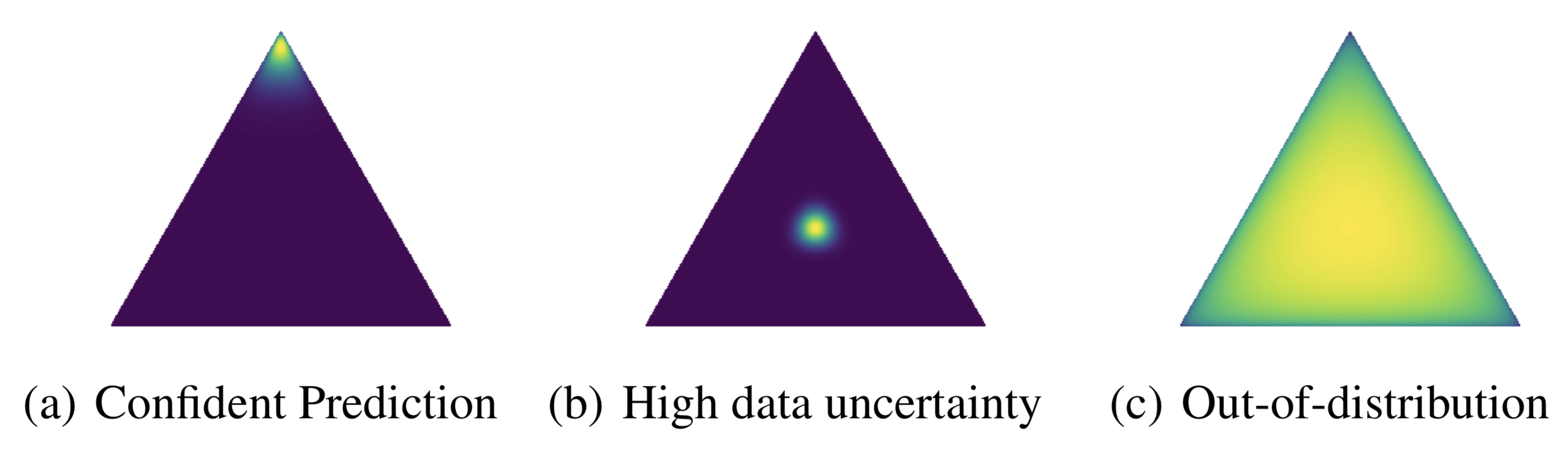}
\caption{An exemplar 3 class simplex taken from a paper by \cite{malinin2018predictive}. This represents the optimal uncertainty behaviour of a classifier given various data points. The lighter colour refers to the probability of a class being placed in the particular area of the space.}
\label{simplex}
\end{figure}

A useful graphical tool in understanding uncertainty can be seen in \cref{simplex}. This figure is known as a simplex and each vertex represents a class in the dataset. This makes a simplex a $K-1$ dimensional object, where $K$ is the number of classes. The particular example shown in \cref{simplex} has been taken from a paper by \cite{malinin2018predictive}. Here there are 3 classes and a lighter colour in this space equates to a higher probability of placing the sample there. Thus, for a confident prediction a vertex will be highlighted like in \cref{simplex}(a), then for uncertain predictions there is \cref{simplex}(b) for an aleatoric case, and \cref{simplex}(c) for the epistemic case. \Cref{simplex} represents the optimal response of a classifier to out-of-distribution errors, as total uncertainty is returned. 

When considering epistemic uncertainty this can again take multiple forms, these are \textit{known} unknowns and \textit{unknown} unknowns. As the names suggest, these are when the types of out-of-distribution data are known beforehand and when it is not. Unknown unknowns present a more challenging problem, as it is much more challenging to prepare the model for something that is unknown. Nevertheless, work in the field can help alleviate these uncertainties, and are predominantly discussed in \cref{epistemic-uncertainties}. Tangentially related fields such anomaly detection and adversarial defence also contribute to the discussion. Anomaly detection models look to identify inputs that do not fit into the distributions on which it were trained and adversarial defence models aim to identify inputs intended to deceive the model. Indeed, a lot of the work in the uncertainty calculation field compares against and shares ideas from these approaches when trying to address the graceful degradation problem. However, both of these methods might leverage information that would not be known. So when considering solutions to the graceful degradation problem it is important to consider which types of unknowns and uncertainties each method is addressing. It is also important to consider the data available to train the system and which supervision strategy is most appropriate during training.

\subsection{Supervision Strategies}

In machine learning there are a spectrum of supervision strategies with which a model can be trained. Supervision in machine learning refers to the error signals with which the model updates itself. In supervised learning a given input has a desired output, that is, it has a label. The learning process can be thought of as like a teacher supervising students. In unsupervised learning the model has no desired output, but rather its goal is to understand the underlying distribution within the data. There also exist intermediate approaches, such as semi-supervised approaches where the model is given a mixture of labelled and unlabelled data, allowing the model to apply \textit{pseudo-labels} to the unlabelled inputs, depending on where it places a point in its distribution. A more recent approach to learning is self-supervised learning, a technique which has massively improved results in the domain of language models, as shown in a paper by \cite{chen2020big}. In the self-supervised paradigm the data itself supplies the label, that is the model is given an altered datapoint and it is asked to reconstruct it. This might be appropriately arranging segments of an image, or selecting a word for a missing part of a sentence. 

Each of these techniques have their own benefits in the representations they allow the model to form. For example, a self-supervised pre-training approach may give a model a better understanding of the underlying data distribution and help it generalise when moving out of this distribution, as is discussed later in \cref{domain-adaptation}. Fully supervised learning largely outperforms other supervision techniques when testing on a single dataset, as shown by the consistent performance of supervised learners in the ImageNet challenge, at the time of writing being led by \cite{foret2020sharpnessaware}. So there are merits for all types of learning, explicit error signals can allow stronger learning of task specific features and a less structured learning process can allow for a more robust inductive bias outside of the training distribution. Hence, all supervision methods will be relevant when models begin to stray from their trained distributions, therefore various supervision strategies are applicable to the graceful degradation problem and will be considered throughout the survey.


\subsection{Taxonomic View of the Problem}\label{taxonomic-view}

\begin{figure}[h!]
\centering
\includegraphics[width=0.98\textwidth]{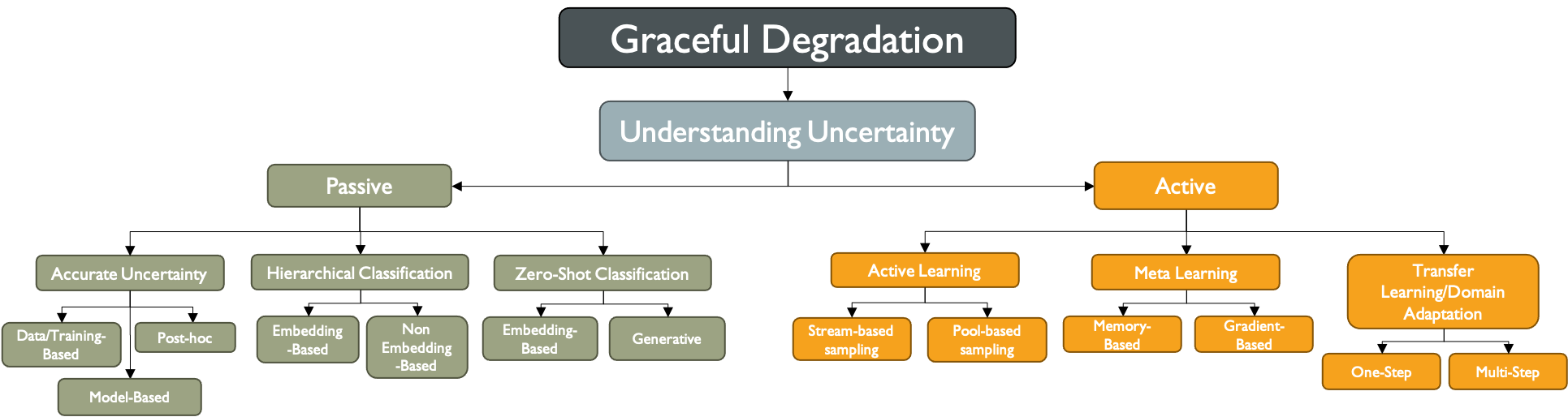}
\caption{A taxonomic view of the graceful degradation problem.}
\label{taxonomy}
\end{figure}

To help understand the different aspects to the graceful degradation problem, it will be beneficial to form a taxonomic view of it. This is shown in \cref{taxonomy}. 

Clearly all of the applications of graceful degradation stem from understanding uncertainty within the input space and are split according to how you deal with that uncertainty. Passive solutions are placed on the left hand side of the diagram and segregated into the three passive approaches to dealing with graceful degradation. Accurate uncertainties refers to the systems that notify an operator of their uncertainty and do not try to function on the data that is causing the uncertainty. This area has been split into three sections that further segregate these approaches, based on where the uncertainty calibration is sourced in relation to the machine learning pipeline. Then hierarchical classification is shown, along with zero-shot classification, which are also split into the different paradigms in the areas.

Active approaches are placed on the right hand side of the diagram, which are further split where appropriate. Meta-learning has been split into the memory-based and gradient-based approaches in few-shot learning. Furthermore, domain adaptation has been split into one-step and multi-step. 

All of the splits in this taxonomy can be segregated further, but for the sake of brevity, these have been omitted from the diagram and will be discussed accordingly in subsequent sections. The remainder of the survey is outlined as follows. First passive approaches will be considered in \cref{passive}, then active approaches will be considered in \cref{active}. Following these, a discussion will be presented in \cref{applications}, which draws comparisons between these two areas and where particular approaches might be the most appropriate. Finally, some conclusions and future work will be presented in \cref{conclusion}.



\section{Passive Approaches}\label{passive}

In the graceful degradation problem passive approaches can be considered as those that once trained are \textit{complete} and require no additional changes to function optimally. Hence, these approaches are self-contained. They can broadly be split into 3 categories when considering the goal of the graceful degradation system and each refer to how they deal with uncertainty. There are those that output accurate uncertainties and express it to the operator without classifying. There are those that will attempt to classify data they are not explicitly trained on. Finally, there are those that will travel up the class hierarchy and offer a coarser-grained classification which they are more confident in. In this section, each approach will be discussed in more detail and notable papers will be highlighted.

\subsection{Accurate Epistemic Uncertainties} \label{epistemic-uncertainties}

As discussed in \cref{uncertainties}, epistemic uncertainty is when the model is uncertain due to a lack of knowledge. That is, the model is uncertain on a classification because the data presented to it is outside of the distribution on which it was trained. Hence, epistemic uncertainties and obtaining accurate estimations of them is very relevant to the graceful degradation problem. This is particularly true when communicating to an operator about the certainty of its output and when it may be necessary for the operator to take control of the classification process. For this reason, methods of obtaining accurate uncertainties will be analysed first.

When considering information outside of the source distribution the type of unknowns need to be considered, for considering known unknowns will prompt different solutions to unknown unknowns. Therefore, throughout the breakdown of tools to calculate epistemic uncertainty, the type of unknowns each tool operates on will be discussed. Out-of-distribution detection models also fit into this category of models as they aim to achieve similar functionality.

In obtaining epistemic uncertainties the field can be broadly split into 3 sections that are defined by the machine learning pipeline: Data/training-based, model-based, and post-hoc. In data/training-based methods you look towards the data in obtaining representations that can produce accurate uncertainties where necessary. In model-based methods the uncertainty framework is built into the architecture of the model itself. Finally, in post-hoc approaches the uncertainty is calibrated by applying the methodology to the output of the model. There also exist hybrid methods, indeed many of the methods at the state-of-the-art employ strategies that utilise ideas from more than one of these sections. However, in segregating and understanding the area these categories logically distribute the approaches in the field and help develop an intuition. 

\subsubsection{Data/Training-based} 

In this category of methods the uncertainties are calibrated using additional data to that which the model is trained on. This might be in a supervised manner or an unsupervised manner (\cite{winkens2020contrastive}, \cite{yu2019unsupervised}). Some papers analyse the distribution of data in the representation space in order to understand predictive uncertainty (\cite{ramalho2019density}, \cite{kamoi2020mahalanobis}). Furthermore, another paper by \cite{kurakin2017adversarial} shows that adversarial examples can persist in the real world even when printed and reprocessed. Hence, data is at the heart of the uncertainty estimation problem, this data can be generated (\cite{Sensoy2018a},\cite{lee2018training}, \cite{lee2018simple}), perturbed (\cite{hafner2020noise}), or it might come from additional datasets (\cite{Li_2020_CVPR}). Here, a leading method which takes the former approach will be discussed. 

\begin{figure}[h!]
\centering
\includegraphics[width=0.35\textwidth]{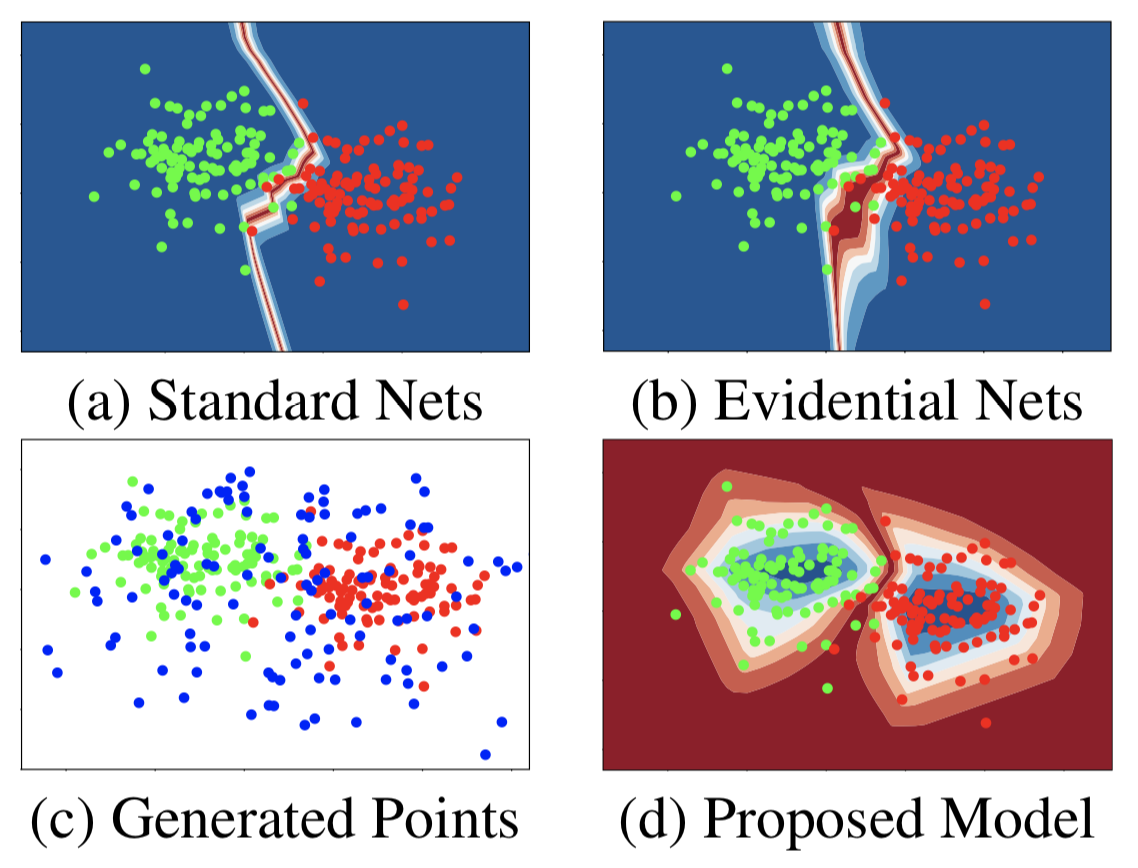}
\caption{A figure taken from the paper by \cite{sensoy2020uncertaintyaware}. In (a) and (b) standard approaches in the field are shown, and the uncertainties are shown near class boundaries. In (c) generated points are presented on which the model can train, and in (d) the class boundary of the proposed model is shown. In (a), (b), and (d) blue refers to highly certain, and red highly uncertain.}
\label{deep-generative-nets}
\end{figure}

\cite{sensoy2020uncertaintyaware} present method that creates uncertainty aware deep classifiers by using generative models to create out-of-distribution samples. The presented approach uses the output of a neural network to model a Dirichlet distribution, representing a probability distribution over possible classes. This distribution is then used to estimate the uncertainty for a given classification. Out-of-distribution samples are then generated using a variational auto-encoder\footnote{A generative model first proposed in a paper by \cite{kingma2013auto}}), the generator being trained to perturb the latent representations for given samples. Two discriminators are trained: One to be fooled, ensuring the representations are sufficiently perturbed; and another which is not fooled, ensuring semantic similarity is preserved. The classification model is then trained on the out-of-distribution samples, as well as the underlying dataset. The effect on class uncertainty is shown in \cref{deep-generative-nets}, where blue and red represent low and high uncertainty, respectively. It is clear the proposed approach (d), recognises the underlying distribution of the classes and takes that into account when classifying data.

The authors evaluate their approach by training it on a dataset with half of the classes held out, it was then tested on a mixture of seen and unseen classes. The key evaluation measure in this procedure are cumulative entropy graphs, namely the area underneath them. Total uncertainty thus high entropy, would be expected on the unseen examples. On seen examples, high entropy should be desired only on the incorrectly labelled samples. This was reflected in their results, which can be seen later in \cref{entropy-classify} and b. This method explores both aleatoric and epistemic uncertainties, hence considering known unknowns and unknown unknowns. Unknown unknowns could be better analysed by passing the model data from a different dataset entirely, such as using fashion-MNIST on a model trained on MNIST\footnote{Modified National Institute of Standards and Technology}, this is something done in a paper by \cite{charpentier2020posterior} discussed in \cref{post-hoc}. However, this is a good example of how extra data can be leveraged in uncertainty estimation and presents a model that is effective in accurately expressing uncertainty in the graceful degradation problem. Moving along the machine learning pipeline, model-based approaches will be discussed. 

\subsubsection{Model-based} 

In model-based approaches the uncertainty is built into the architecture of the model and is generated intrinsically. Such approaches can take distributions over the model parameters like in bayesian neural networks, or they might leverage other information such as that from gradients to evaluate uncertainty. 

In bayesian frameworks the uncertainty between the training and test distribution are considered to arise from uncertainties in the model itself. To instantiate one of these systems using a neural network you apply a probability distribution over all of the parameters of the network, hence removing the deterministic nature of a forward pass. The inference process in bayesian neural networks involves a number of forward passes, thus allowing the weight uncertainties to propagate through the network and giving a probability distribution for the output. These can be sampled in a variety of ways the most popular of which being variational inference, a technique popularised by \cite{blundell2015weight} where \textit{Bayes by Backprop} was introduced. This method would catalyse the development of the variational approach as it showed backpropagation could work as intended when training these probabilistic networks, something which previously presented a significant challenge to the field.\footnote{Monte carlo methods are also prevalent in the field, but these are more computationally demanding, and will not be discussed.}

Lots of the work in the bayesian network space has expanded on this, such as obtaining weight uncertainties implicitly using \textit{hypernetworks} \cite{pawlowski2018implicit}, or by viewing dropout as a form of bayesian inference \cite{gal2016dropout}. Another paper by \cite{tran2019bayesian} restricts the bayesian properties to single layers to analyse uncertainty. Here we will highlight the hypernetwork approach. However, the reader can find an implementation orientated review of bayesian approaches in a recent survey by \cite{jospin2020hands} and a more comprehensive view on uncertainty quantification in another by \cite{abdar2020review}. 

\begin{figure}[h!]
\hspace{\fill} 
  \subfloat[]{%
  \label{bbh-diagram}
   \includegraphics[width=0.48\textwidth]{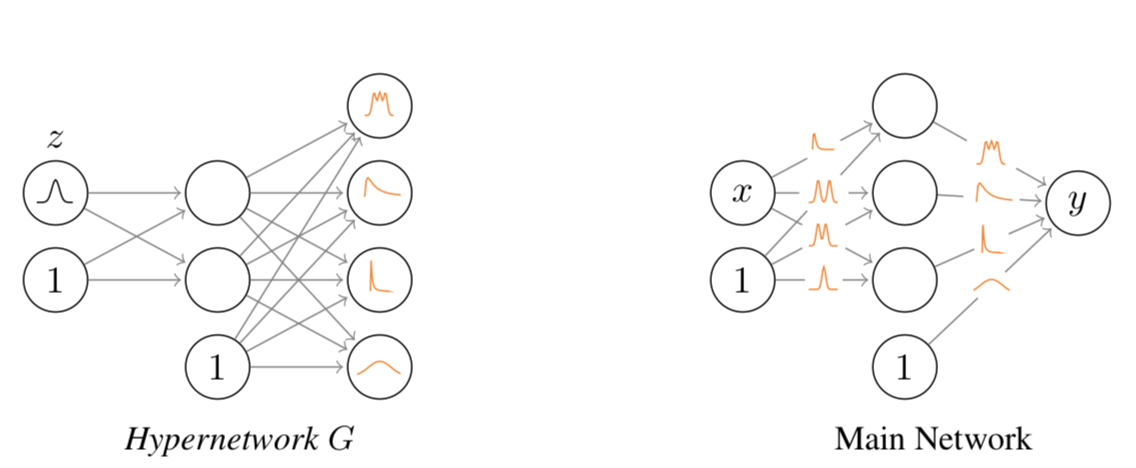}}
\hspace{\fill}
  \subfloat[]{%
    \label{bbh-figure}
   \includegraphics[width=0.45\textwidth]{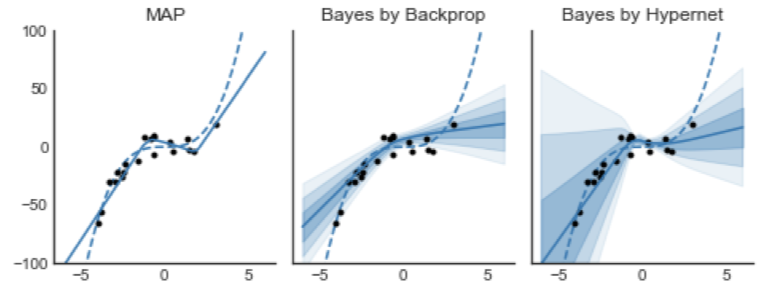}}
   \hspace{\fill}
  \caption{In (a) the interaction between the hypernetwork and the main network is shown, with the hypernetwork producing the weight distributions for the second layer of the main network. In (b) the effect on model uncertainty is shown on a regression task for 3 different models. Maximum a posteriori (MAP), \textit{Bayes by Backprop} (\cite{blundell2015weight}), and the hypernet approach (\cite{pawlowski2018implicit}) here the lighter shade of blue refers to a higher uncertainty, the blue dashed line refers to the ground truth, and the solid line the prediction of the model.}
\end{figure}

Hypernetworks, first presented in a paper of the same name by \cite{ha2016hypernetworks}, are a method of using a neural network to generate the weights for a target neural network. \cite{pawlowski2018implicit} utilise hypernetworks to obtain implicit distributions of weights, such distributions have intractable probability densities but allow for very easy sampling. The authors sample from the weight distributions given by the hypernetworks during a forward pass. This is reflected in \cref{bbh-diagram}, where the hypernetwork $G$ produces a weight distribution for the second layer of weights in the main network. The predictive uncertainty in the model is also reflected using a toy regression task, shown in \cref{bbh-figure}. This shows their method exhibits the best trade-off between regression accuracy and predictive uncertainty, when compared to other methods in the field. 

To train this system the hypernetwork generating the weights needs to be optimised, they do this through a method that closely resembles adversarial training. A generator models the variational distribution of the main network, a discriminator then approximates a density ratio between model parameters given those of the generator, and the probability of generator parameters. The authors show this information allows them to effectively approximate the evidence lower bound (ELBO), something used to optimise bayesian neural networks in the field.

\begin{figure}[h!]
\hspace{\fill} 
  \subfloat[]{%
  \label{entropy-classify}
   \includegraphics[width=0.45\textwidth]{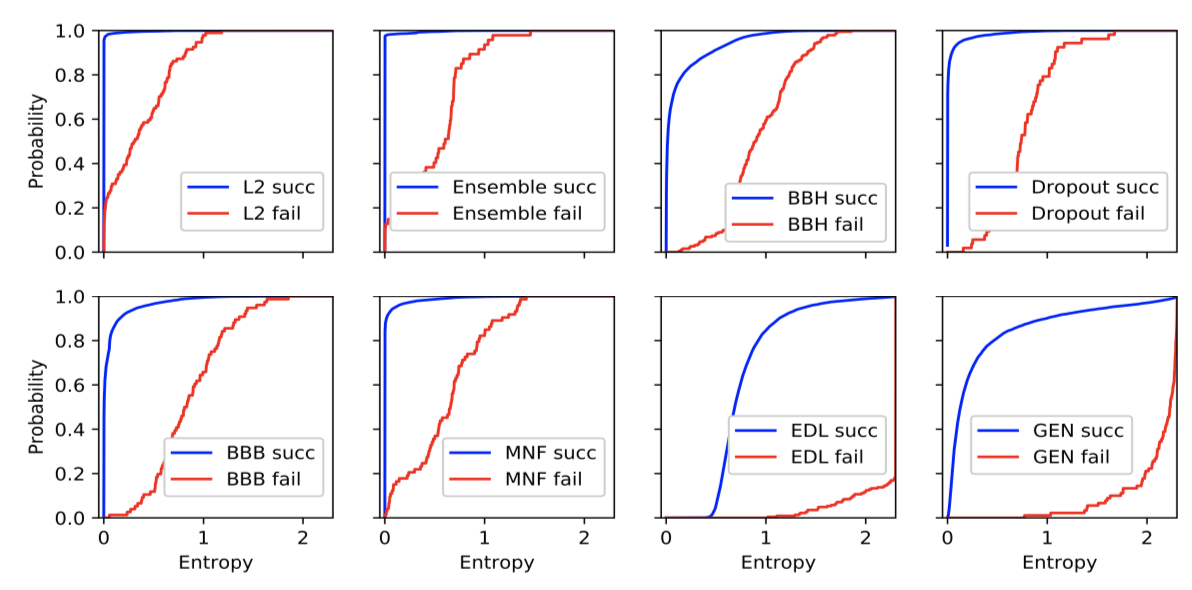}}
\hspace{\fill}
  \subfloat[]{%
  \label{entropy-ood}
   \includegraphics[width=0.35\textwidth]{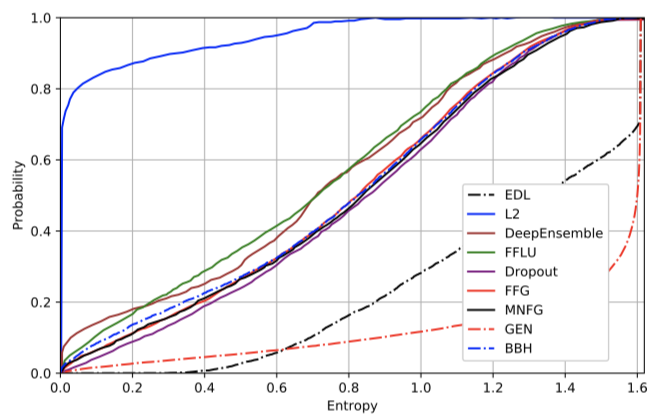}}
   \hspace{\fill}
  \caption{Two figures taken from the paper by \cite{sensoy2020uncertaintyaware}, showing cumulative density plots of entropy. In (a) correct and incorrect distributions are compared are compared for various uncertainty aware approaches. In (b) the distributions are shown for when they encounter out-of-distribution data. In both \cite{sensoy2020uncertaintyaware} and  \cite{pawlowski2018implicit} the area underneath these plots is used as a form of analysis.}
\end{figure}

As part of the analysis the main network is tested on in-distribution images and out-of-distribution images to analyse the generalisability of the method and its uncertainty awareness. The performance is quantified on these distributions by using the area under cumulative density plots of entropy. This is the same analysis as that in the paper by \cite{sensoy2020uncertaintyaware}, examples of which are shown in \cref{entropy-classify} and b. Here it is named the area under curve (AUC), not to be confused with the area under the receiver operator curve, which is the area under the curve of true positive rate against false positive rate and generally called the AUC. 

The argument for using this form of analysis is that if the model is confident in its prediction, as in the in-distribution case, the AUC should be high as the entropy will more frequently be a low value, shown in \cref{entropy-classify}. Whereas in the low confidence out-of-distribution case, the entropy should be low until close to the maximum entropy value, shown in \cref{entropy-ood}. The method outperforms all other competing methods in both maximising the in-distribution AUC and minimising that on out-of-distribution data. 

The key advantage of the method is its flexibility and scalability, as the use of hypernetworks to generate the weights implicitly accounts for the architectural complexity of the main network. This allows the method to scale to very large and complex networks, something intractable in competing bayesian approaches. Hence, if the underlying classification problem is one that requires the representational capacity of a larger, more complex network, this method will suitably provide uncertainty measures for the problem. However, the authors only consider known unknowns in their analysis. That is, data semantically similar to that which it was trained on. Hence, this method may not be suitable when the model is likely to encounter data far outside of the training domain: Unknown unknowns. The authors could have undertaken such analysis, however, the authors do test their method on adversarial samples and show it performs better than competing bayesian approaches. Another recent approach also uses hypernetworks when identifying out-of-distribution samples, which is presented in a paper by \cite{ratzlaff2020hypergan}.  

There have been a number of papers that use non-bayesian approaches to this problem. Here the uncertainty can be calculated using signals from the model itself, this might be gradient information, as in a paper by \cite{oberdiek2018classification}, or using ensembles as in a paper by \cite{kachman2019novel}. Furthermore, a paper by \cite{Hein_2019_CVPR} analyses the ReLU activation function in neural network design, proving it can aggravate out-of-distribution errors. An interesting non-bayesian approach is presented in a paper by \cite{geifman2018biasreduced}, where the authors suggest highly confident predictions are an artefact of dynamic and stochastic training processes. The paper suggests uncertainty signals obtained from models which are optimised on a classification problem will be inherently biased towards making predictions with high confidence. It is argued this is a form of overfitting which prompts sub-optimal behaviour on out-of-distribution samples and that the stochastic gradient descent process deforms reliable confidence estimates throughout the training process. 

To address this, more accurate readings of predictive uncertainty can be obtained from earlier \textit{snapshots} of the network. To do this, weights from earlier iterations of the network are sampled to give accurate estimates for uncertainty. Two methods of obtaining these weights are proposed. One trains an input-wise selection mechanism to identify earlier iterations of the model which improve the uncertainty estimation, this method requiring an auxiliary dataset. The second method looks to approximate the first method without the use of an auxiliary set, this has the benefit of shorter training periods in the model at the cost of longer inference times. This second method, which leverages the fact that easier samples are learned earlier on in the training process, performs surprisingly well and is used for the majority of their analysis. The method outperformed all competing methods at the time, when maximising the confidence score of correct predictions. The analysis fails to include out-of-distribution samples, hence only aleatoric uncertainty is considered. However, the method performs well and is a good example of a model-based approach to the uncertainty estimation problem, if not explicitly used in the graceful degradation setting. Now the focus will be shifted towards the end of the machine learning pipeline in post-hoc methods.

\subsubsection{Post-hoc}\label{post-hoc}

In post-hoc approaches you focus on the output of the model and use it to calibrate the predictive uncertainty. The benefit of these approaches is that they can often be incorporated with any architecture and arguably any machine learning model. These methods look to transform the output so that it more accurately reflects the confidence of the prediction, this might be through simple temperature scaling or use more complicated measures like in a recent paper by \cite{kull2019beyond} which utilises Dirichlet distributions. Other recent methods include knowledge distillation of uncertainties (\cite{pouransari2020extracurricular}), utilising likelihood ratios (\cite{Ren2019}), or minimising KL divergence between OOD samples and a uniform distribution (\cite{lee2018training}). 
 
\begin{figure}[h!]
\centering
\includegraphics[width=0.8\textwidth]{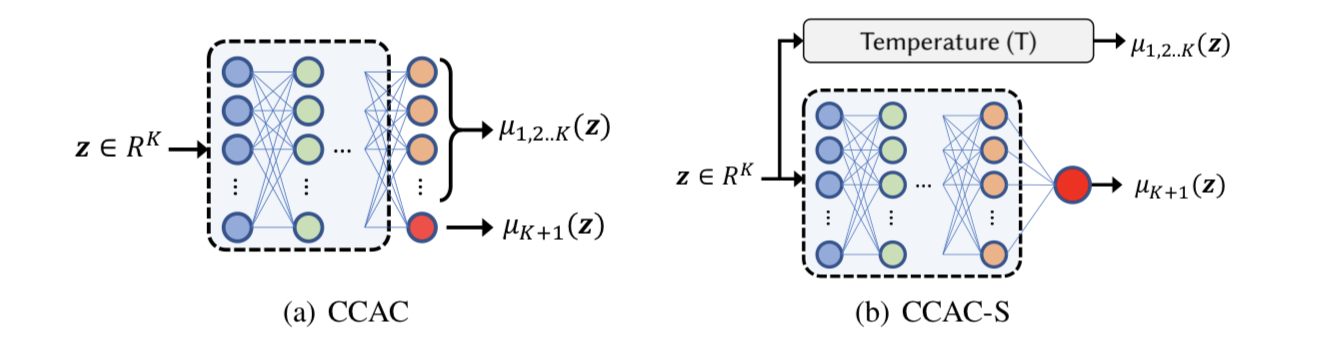}
\caption{A figure from the paper by \cite{shao2020calibrating} the authors depict their two approaches, with the soft-max logits $\bf{z}$ being taken as an input, and the approach outputs a calibrated soft-max. The two approaches are compared, the complete version in (a) and the simplified version in (b).}
\label{CCAC}
\end{figure}

A particularly relevant post-hoc approach explicitly calibrates their model on out-of-distribution datasets, presented in a recent paper by \cite{shao2020calibrating}. The authors incorporate an auxiliary classifier to the post-hoc framework that identifies mis-classification. The classifier is a simple neural network which maps the model output to a calibrated soft-max, which can be used to give more appropriate confidence estimates. An auxiliary class is also introduced which identifies miss-classified samples and is used to calculate the confidence of a given classification. Two strategies of doing this are developed, one where the auxiliary classifier has an output neuron for every class and an additional one for the auxiliary class. The second method simply temperature scales the output of the model and the auxiliary classifier has a single output which predicts the classification accuracy. These are shown in \cref{CCAC}a and \cref{CCAC}b respectively in a figure taken from the paper. The auxiliary classifier was trained in a supervised manner, aiming to map the class of incorrect classifications to the auxiliary class.

The models were trained on the CIFAR-10\footnote{Canadian Institute For Advanced Research} dataset and tested using a modified CIFAR-10 dataset, which was synthesised using CIFAR-100 classes. It included some some semantically similar classes from CIFAR-100 which were mapped to the similar class in the CIFAR-10 set, representing known unknowns. Unknown unknowns were analysed by including completely out-of-distribution samples, represented using classes that were not semantically similar. The method was able to accurately identify known samples with high confidence, record low confidence in incorrect predictions, and would consistently use the auxiliary class when encountering out-of-distribution data. Hence, this is very relevant in the graceful degradation problem in that the model performs optimally when leaving the distribution on which it was trained. This is shown both close to the distribution and well outside of the distribution. The analysis shown in this paper is perhaps the best and most practical analysis to perform when analysing graceful degradation problems. The method does necessitate the training of the auxiliary classifier which can be expensive, but there it is very flexible regarding the architectures on which it can be used.

\begin{figure}[h!]
\centering
\includegraphics[width=0.4\textwidth]{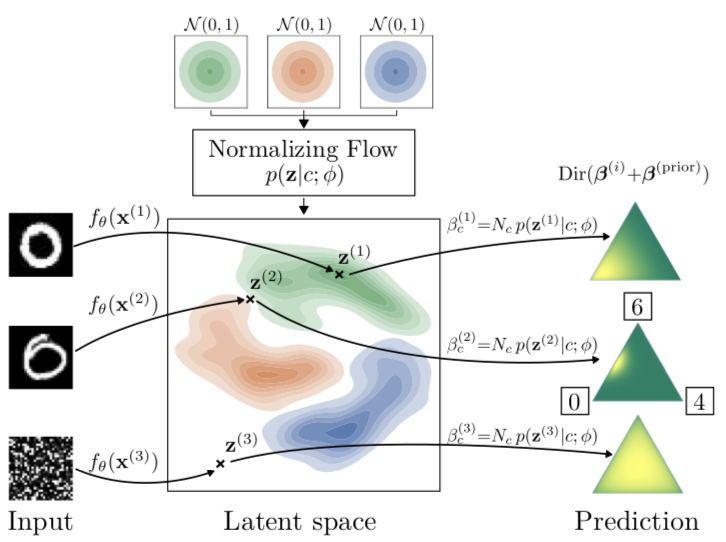}
\caption{A graphical representation of the model in the paper by \cite{charpentier2020posterior} taken from the same paper. The inputs are mapped to a latent space, the densities within this space are morphed by normalising flows. Predictions for various inputs are sampled from this space and shown in simplexes. For the sake of intuition, only 3 classes are shown in the simplex examples.}
\label{postnet}
\end{figure}

Another similar method at the state-of-the-art is \textit{Posterior Networks} proposed in a paper by \cite{charpentier2020posterior}. The authors propose a neural network which uses normalising flows to predict an individual closed-from posterior distribution over probabilities for a given input sample. It is comprised of 3 components, an encoder which maps inputs to a latent position, a normalising flow which provides density estimations for each class in the latent space, and a bayesian loss which allows for uncertainty aware training. This is shown in \cref{postnet}. The key problem solved in this paper is that the method is designed to work without the knowledge of known unknowns and does not leverage any additional data.

The method is evaluated on in-distribution samples and out-of-distribution samples for both known unknowns and unknown unknowns. This can be also seen in \cref{postnet}. Furthermore the predictive uncertainties are reflected in the simplex diagrams to the right hand side, matching those in \cref{simplex}. The training datasets included CIFAR-10 and MNIST. The predictive uncertainty was evaluated on unseen datasets, left-out classes, out-of-domain, and on dataset shifts. State-of-the-art results were achieved in all areas and is very relevant and successful in the graceful degradation problem of communicating uncertainty. Rather than expressing uncertainty upon leaving the data distribution, it may be more desirable for the model to function in the encountered domain instead, this is where zero-shot learning could provide insights.

\subsection{Zero-Shot Classification}\label{zero-shot}

In the field of zero-shot learning the goal is to design models that can classify data it has never seen explicit examples of. Hence, this area is about the model understanding the space outside of what it knows and performing well in this area. This technique clearly only works on known unknowns, but it will be beneficial to discuss zero-shot learning techniques in the context of the graceful degradation where known unknowns exist.

In zero-shot classification there exist two main protocols: Conventional zero-shot learning and generalised zero-shot learning (GZSL). These methods differ in their targeted test distributions. In conventional zero-shot learning the test distribution is only the data on which zero-shot classification should be achieved. In GZSL the test set will include both classes from the training distribution and novel classes on which the model should achieve zero-shot classification. The latter is arguably the most relevant in the graceful degradation problem, as a gracefully degrading system should also perform optimally on the trained distribution. Hence, some papers from this area will be considered. Zero-shot learning has also been used in other vision tasks, namely semantic segmentation, recently shown in a paper by \cite{NIPS2019_8338}. 

In GZSL the goal is often to define a shared embedding space on which the seen and unseen classes can develop discriminatory representations. The embeddings for the unseen classes are often formed using additional semantic information. These might be in the form of attributes (\cite{farhadi2009describing}, \cite{liu2020hyperbolic}, \cite{huynh2020fine}), word embeddings (\cite{mikolov2013efficient}, \cite{norouzi2014zeroshot}, \cite{sariyildiz2019gradient}), textual descriptions (\cite{reed2016learning}, \cite{elhoseiny2019creativity}), or might employ generative techniques (\cite{yu2020episodebased}, \cite{Han_2020_CVPR}, \cite{ni2019dual}, \cite{sariyildiz2019gradient}). One recent implementation in a paper by \cite{wu2020self} uses self-supervision to learn domain-aware features in the GZSL task, something that may also be relevant in domain adaptation which is discussed in \cref{domain-adaptation}. As an example, GZSL systems be applicable in drone systems that need to classify objects which only have semantic descriptions, like in the first vignette.

\begin{figure}[h!]
\hspace{\fill} 
  \subfloat[]{%
  \label{over-complete}
   \includegraphics[width=0.25\textwidth]{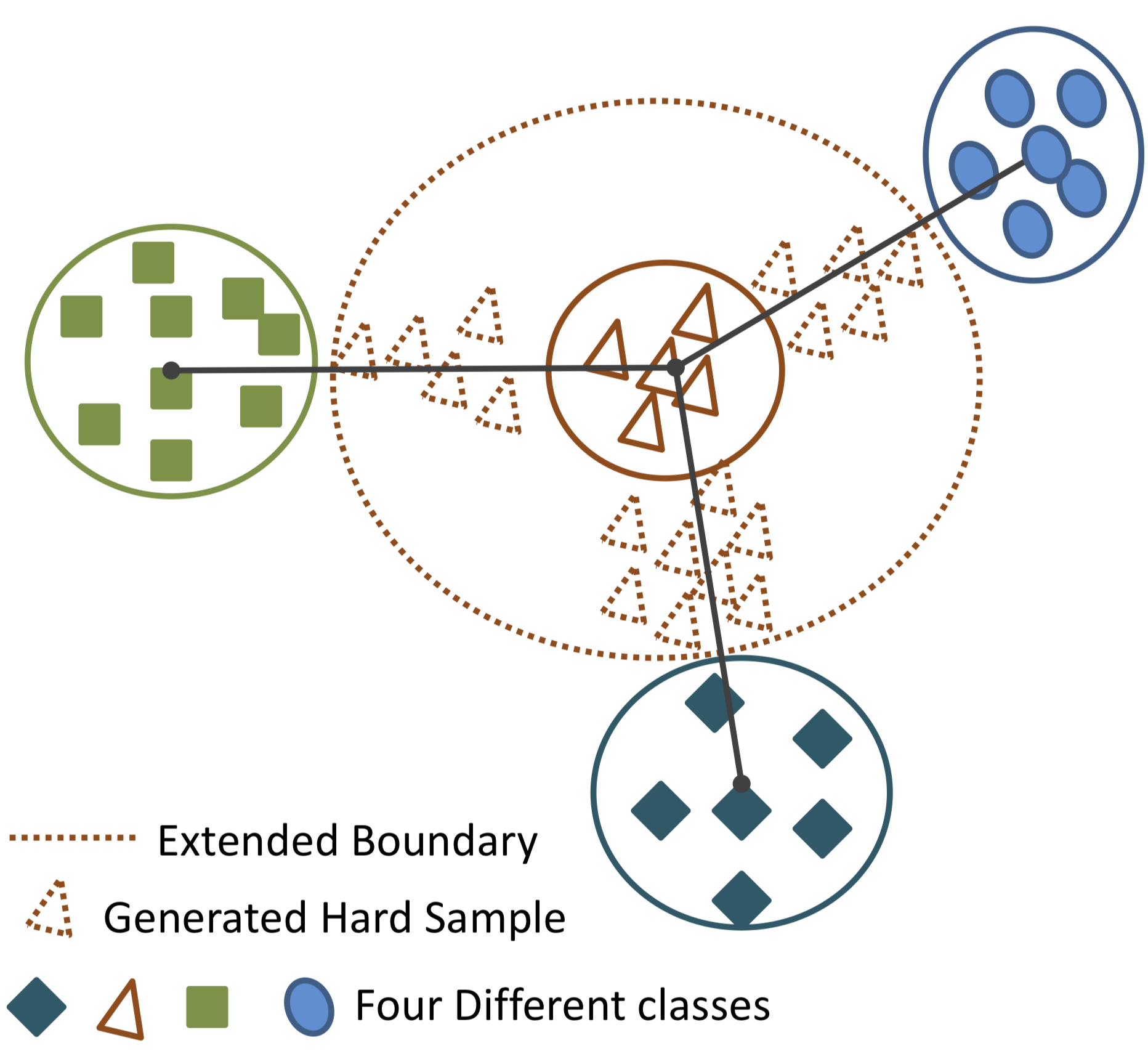}}
\hspace{\fill}
  \subfloat[]{%
    \label{over-complete-pipeline}
   \includegraphics[width=0.6\textwidth]{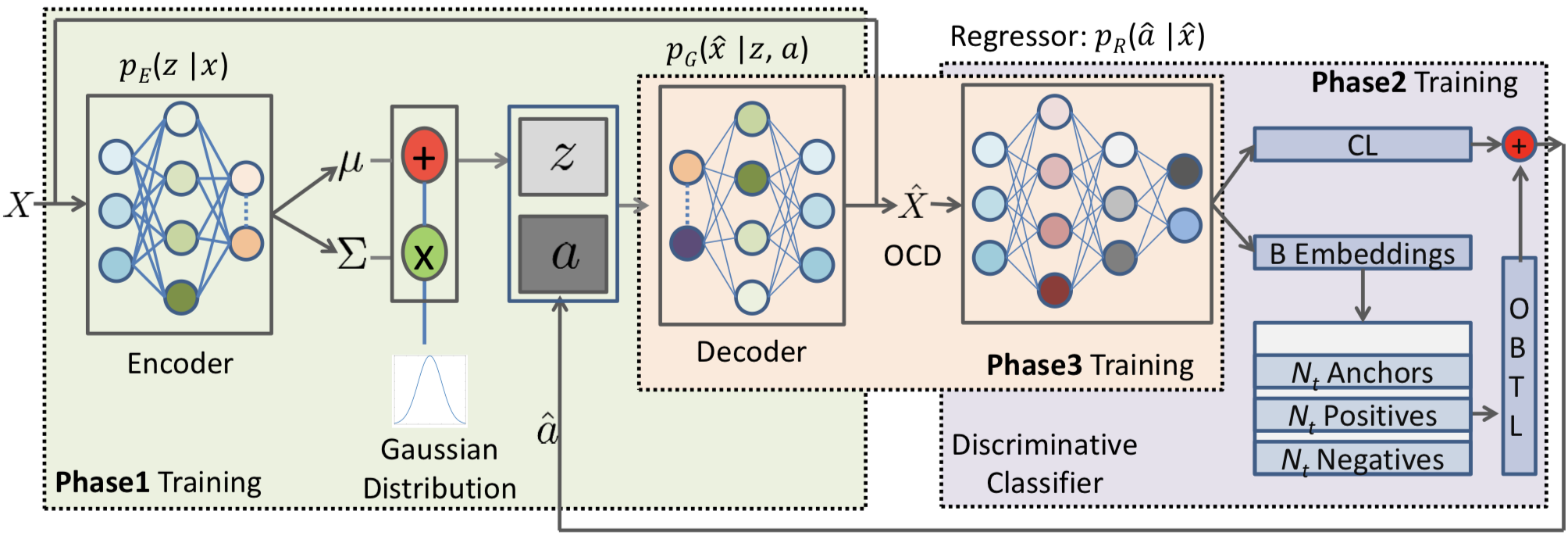}}
   \hspace{\fill}
  \caption{Two figures taken from the paper by \cite{keshari2020generalized}. In (a) the process of generating over-complete samples is shown with relation to the class space, and how it creates an extended boundary by producing hard samples. In (b) the model structure is shown in its various components and how each of those relate to phase of training.}
\end{figure}

The authors of \cite{keshari2020generalized} propose that one of the primary reasons GZSL systems fail is because they are required to classify \textit{hard} samples of unseen classes at test time. That is, the samples are close to another class, thus causing confusion in the classifier. To that end, the authors of the paper propose a new attribute based methodology of training GZSL systems, whereby they train the model given an \textit{over-complete} distribution of samples. In the proposed methodology, an over-complete distribution is one that aims to improve the generalisability with the respect to the unseen classes. This is done by generating examples of the unseen class that are close to the distributions of seen classes, hence producing an \textit{extended boundary} for the unseen class. This allows the model to generalise better when seeing the real instances of the class. The resultant effect on class embeddings can be seen in \cref{over-complete}, where all possible hard samples are generated relative to each class. 

\Cref{over-complete-pipeline} demonstrates the structure of the approach. The model works by first mapping inputs $x$ to a latent position $z$ using an encoder, a decoder then generates samples $\hat{x}$ given the latent encoding $z$ and attributes $a$. The encoder and decoder combination are known as a conditional variational auto-encoder (CVAE), in this instance it is conditioned on the attributes of the inputs. Finally, a regressor maps the samples $\hat{x}$ to their predicted attributes $\hat{a}$ and is optimised using an online batch triplet loss (OBTL) and a centre loss (CL). The OBTL ensures high inter-class separability and low intra-class separability, and the CL ensures that the generated points for a given class keep the distribution centred about its learned centre. 

The model is trained in 3 phases, which can also be seen in \cref{over-complete-pipeline}. In the first phase the CVAE is trained on seen data to produce a latent position $z$ given $x$, which is subsequently appended to the attributes of the data point, $a$, creating a vector $[z,a]$. This is used to create the generated samples $\hat{x}$ and is trained using KL divergence and conditional marginal likelihood. The trained CVAE should be able to generate synthetic data given attributes $a$. In the second phase of training the regressor/classifier is trained using the OBTL and CL to optimally map synthetic data to corresponding attributes. The third phase of training regularises the over-complete distribution to ensure it creates class specific examples even when randomly sampling from the embedding space. The authors find their model would outperform most existing approaches on common ZSL datasets: Animals with Attributes 2 (AWA2)(\cite{lampert2013attribute}), Caltech-UCSD Birds (CUB) (\cite{wah2011caltech}), and SUN (Scene UNderstanding) (\cite{patterson2012sun}). 

\begin{figure}[h!]
\hspace{\fill} 
  \subfloat[]{%
  \label{hyperbolic-diagram}
   \includegraphics[width=0.55\textwidth]{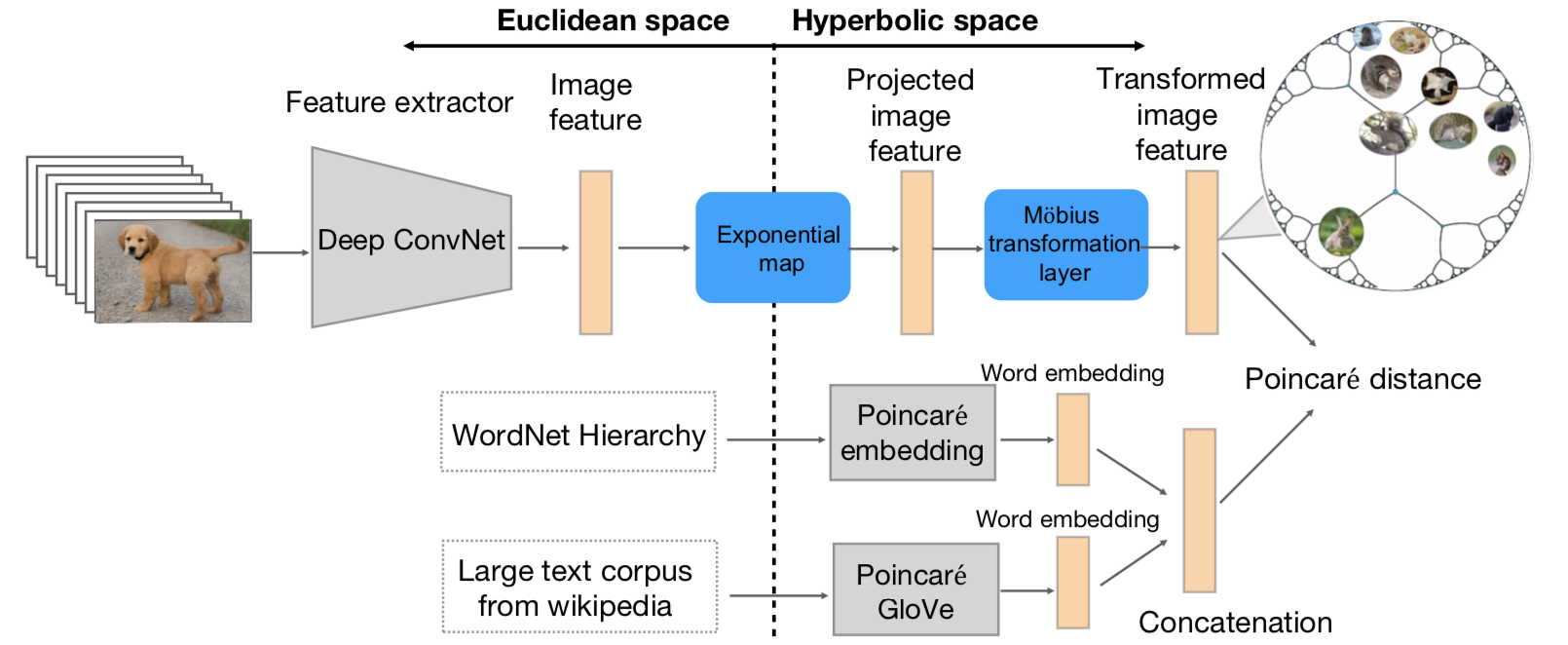}}
\hspace{\fill}
  \subfloat[]{%
    \label{hyperbolic-hierarchy}
   \includegraphics[width=0.35\textwidth]{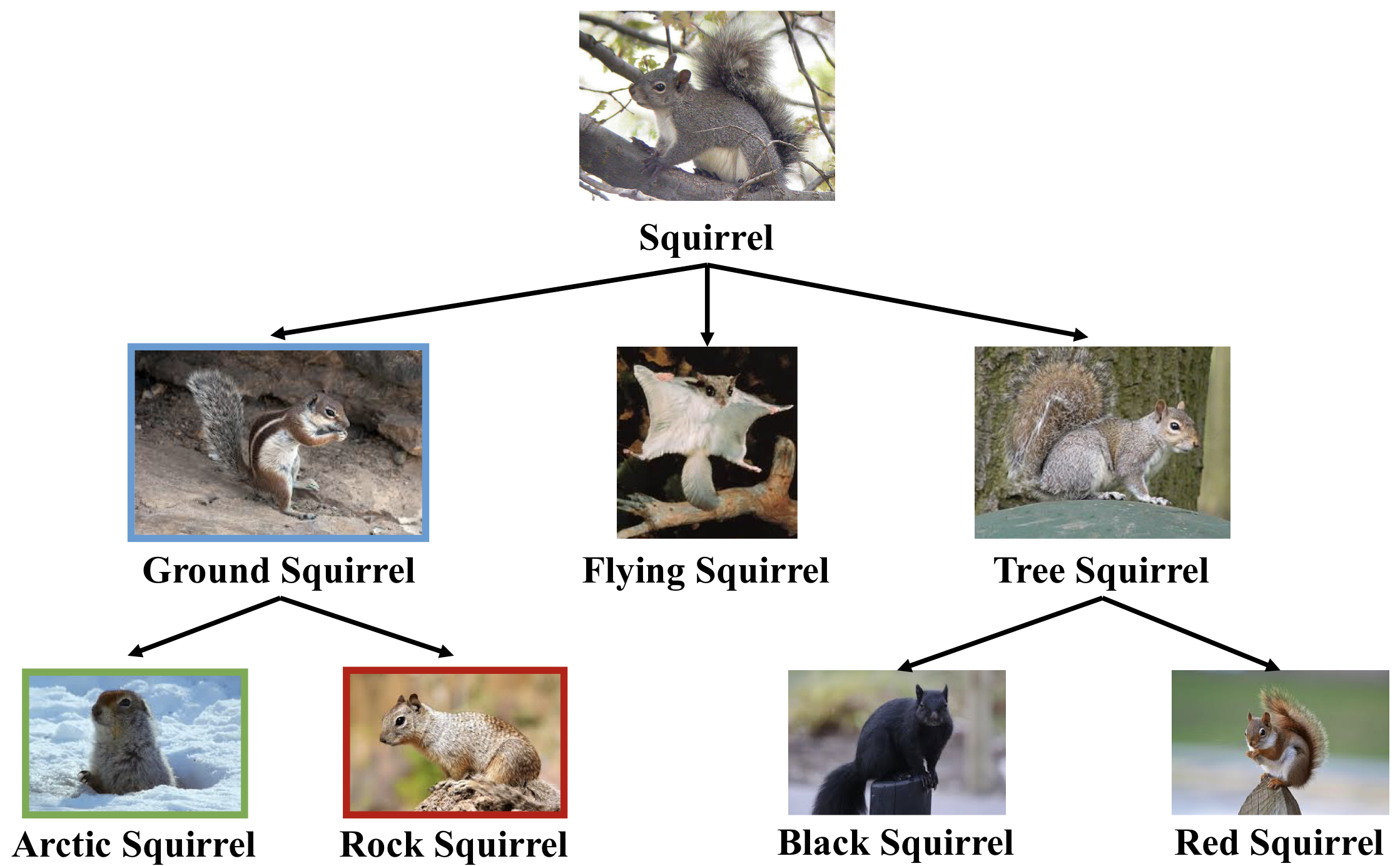}}
   \hspace{\fill}
  \caption{Two figures taken from the paper by \cite{liu2020hyperbolic}. In (a) the structure of the proposed approach is presented. In (b) the hierarchy of a given class is shown, in this case squirrels.}
\end{figure}

A recent implementation by \cite{liu2020hyperbolic} approaches the zero-shot learning problem in a way that reflects the hierarchy of classes and uses word embeddings to do so. Zero-shot learning techniques typically use euclidean embedding spaces, in this paper the authors project each sample to a hyperbolic embedding space, the first example of such an approach in zero-shot learning. The embedding space used is the same as that in a paper by \cite{Nickel2017}, a Poincar\'e hyperbolic space. This allows hierarchical representations in the input space and the distances within it are differentiable. The framework of their approach has two components, shown in \cref{hyperbolic-diagram}, a figure taken from the paper. 

The first of these components is the learning of a hyperbolic label embedding which embeds image labels into a hyperbolic space, encoding hierarchical information as well as semantic information in doing so. Poincar\'e embeddings are used to encode WordNet noun hierarchies into what is known as a Poincar\'e ball, a disk in Poincar\'e space where the distance scales exponentially as the edge of the disk is approached. This prevents hierarchical representations from becoming increasingly clustered as the edges are made, allowing the embeddings to learn the hierarchical connections of WordNet whilst preserving the semantical distances between the words themselves. The authors also employ GloVe word embeddings to capture semantical relationships between classes, this also being trained in a Poincar\'e ball space. This produces two embeddings which are concatenated to form a single embedding for image labels which has structural and semantical information encoded within it. 

The second component in their approach is the learning of image embeddings, this process consists of two components: An exponential map, which projects image features into the hyperbolic space; and a transformation network which learns to align the new image representations with their respective labels in the hyperbolic space. The resultant system is one that transforms image embeddings into a shared hyperbolic space with the word embeddings, meaning both hierarchical and semantic relationships can be learnt in visual representations. Such a hierarchy can be seen in \cref{hyperbolic-hierarchy}, another figure from the paper. The model was trained to minimise the distance between the image embeddings and the label embeddings.

Experiments were conducted on ImageNet (\cite{deng2009imagenet}), a popular benchmark in the ZSL community. The unseen classes are segregated by the number of connections they are away from a seen class, making 3 different test sets: 2-hop, 3-hop, and All. The model is evaluated on all three of these test sets, both in the ZSL and GZSL paradigms. The model, whilst not achieving perfect top-1 accuracy in classifying unseen classes, would correctly output the parent class in many cases. That is, the model would often include a label further up the class hierarchy as opposed to another fine grained prediction. Arguably this ability to predict the immediate parent class of a model reflects robustness in a ZSL model, as it shows an understanding of the class space. Furthermore, such hierarchical predictions make systems more applicable in real world scenarios, particularly in the graceful degradation setting where an operator would prefer a confident coarse-level prediction, than a single low confidence fine-level label. Indeed, such approaches will be discussed in the next section. 

Zero-shot learning approaches are relevant in the graceful degradation problem as the models aim to understand the space outside of the seen classes and looks to be better prepared as they approach these distributions. These models necessitate the existence of known unknowns, limiting their applicability. However, there are an abundance of datasets that can help facilitate the training of such systems. Although it is not explicitly discussed, it is clear such systems overcome epistemic uncertainty in making predictions, as they classify data outside of the training distribution. Although, zero-shot learning techniques often lack reliable confidence estimates and as such will not be applicable to fully autonomous systems where high error rates are not permitted, let alone over-confident errors. A way around the high error rate of these systems is to produce hierarchical outputs, like in the previous approach. More explicit investigations of hierarchical classification have been undertaken in the field and will be discussed in the next section.

\subsection{Hierarchical Classification} \label{hierarchical-classification}

In the hierarchical classification problem, rather than asking the model to make a prediction in an area in which it is uncertain, the model is instead asked to predict a parent class at a level of hierarchy at which the it is more confident. For example, if the model failed to recognise an animal as a pigeon it might be able to recognise that the animal is a bird and offer a range of classes the bird might fit into. This can be seen as a compromise between the other passive approaches to the graceful degradation problem: The image is not classified due to uncertainties, but more information is offered as opposed to simply returning an uncertainty. 

Like in the aforementioned paper by \cite{liu2020hyperbolic}, recent investigations into hyperbolic embedding spaces (\cite{Khrulkov_2020_CVPR}, \cite{Chami2019}) have proven to be particularly insightful in the hierarchical labelling problem (\cite{lopez2020fully}, \cite{dhall2020hierarchical}). However, such investigations have been more recent and there is a wealth of existing literature in hierarchical image classification that takes more conventional approaches. These can broadly be split into embedding-based approaches (\cite{faghri2017vse++}, \cite{lagrassa2020learn}, \cite{Seo2019}, \cite{koo2019combined}, \cite{guo2018cnn}) and non-embedding based approaches (\cite{Bertinetto_2020_CVPR}, \cite{8260647}, \cite{kumar2017hierarchical}, \cite{chen2018fine}, \cite{deng2014large}), which typically exploit label hierarchies. Another interesting approach by \cite{cha2020hierarchical} takes the superclass label as an input to help the fine-grained classification. Furthermore, hierarchical labels were also implemented in an active learning paper by \cite{nakano2020active}. Active learning approaches will be considered in \cref{active-continual}, but this shows the relationships between the various avenues of the graceful degradation problem. To give context, hierarchical labels might be used in a disaster response system where it cannot classify a particular object, but it might be able to classify it as potentially endangering to, or holding, survivors .

A particularly relevant technique in this area is presented in a paper by \cite{achddou2020nested}, where a new type of learning is proposed, \textit{nested learning}. In this paradigm the model should understand the hierarchical nature of the input and provide information of varying specificity depending on the data sample. This allows the model to output confidence levels for classes at different levels of granularity, including the fine-grained class. The core work in the paper discusses and formulates such learning problems and it is shown that systems trained in such a manner can offer meaningful coarse-level predictions even when the fine-level predictions are severely degraded.

\begin{figure}[h!]
\centering
\includegraphics[width=0.75\textwidth]{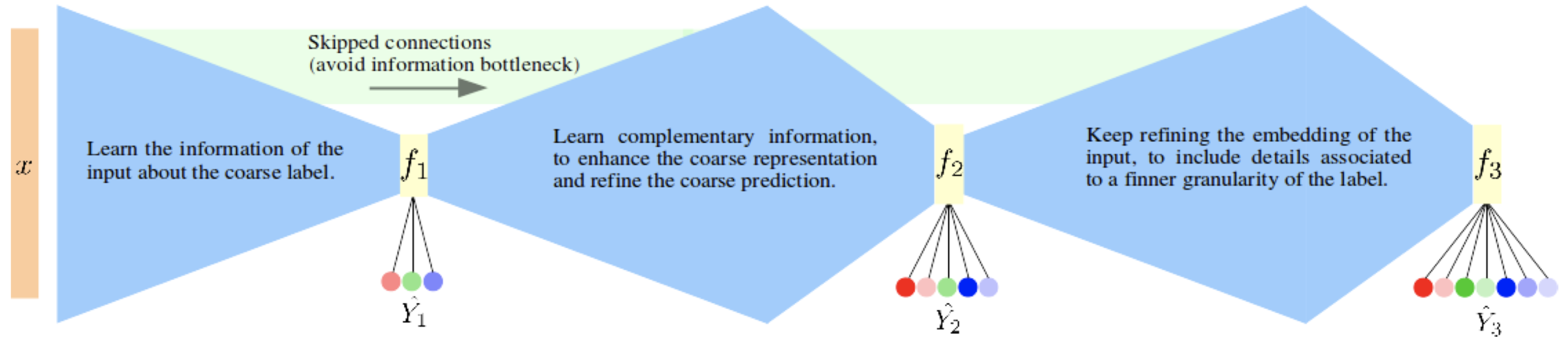}
\caption{A figure taken from the paper by \cite{achddou2020nested} presenting the structure of the approach.}
\label{nested-representations}
\end{figure}

Four main components are designed to make such systems work: A sequence of low dimensional nested representations, calibrated output predictions, the combination of nested predictions, and a practical and stable training protocol. The process of extracting nested representations is shown in \cref{nested-representations} where by the input $X$ learns information about the coarse label $\hat{Y}$ through the classification using $f_1$ and uses this to inform the following label at the next classification layer. Between these classification layers the input is compressed sequentially but skipped connections also allow information to flow between the layers. The second step involves learning complementary information that when combined with output of a coarse classification layer, $f_{n}$, can provide meaningful insights to classification at $f_{n+1}$. Here the skipped connections are necessary to prevent the input from being degraded through when coarse level predictions. The score outputs need to be calibrated as a soft-max of a given class at layer $f_n$ does not necessarily equate directly to its probability. This is because at a given coarse-level classification, the number of classes available is no longer equal to all of the classes across all layers, hence they need to be re-normalised. Finally, the authors train their model in a cascaded manner, gradually adding finer level classes iteratively as the coarser predictions are optimised, thus optimising deeper parameters as training goes on.

To evaluate their method a number of datasets were used to train the models, including Fashion-MNIST, CIFAR-10, and CIFAR-100. The latter dataset has an accompanied taxonomy, the former had such taxonomies manufactured by the authors. Results suggest that coarsely annotated data would allow for more robust and less over-confident fine-level predictions. Furthermore, given a specific training budget, the proposed approach still outperforms a typical end-to-end approach. When given adversarial examples the approach still showed robustness and a reduced over-confidence. The authors fail to include state-of-the-art papers in their comparison, opting instead for an end to end version of the same architecture, whilst this does highlight the benefits of the approach, it does not show the effectiveness of it in the context of the field.  Something also neglected in the analysis is out-of-distribution samples, whether or not the nearest parent label can be identified by such models would be an interesting analysis. So whilst the authors do not explicitly investigate the graceful degradation problem, they do investigate tangentially related areas such as adversarial defence and uncertainty in classification. Furthermore, the paper does suggest an uncertainty-centric method, that provides such estimations at varying levels in the class hierarchy. This is something desirable in graceful degradation, albeit in this instance limited to the aleatoric case. Furthermore, it is clear such methods compliment branched neural network structures due to the hierarchical nature of the output branches, indeed such structures have recently been shown to allow for greater adversarial robustness, efficiency, and accuracy in a recent publication by \cite{hu2020triple}.

Hierarchical methods such as these could prove promising in the graceful degradation field, a model understanding uncertainty throughout the class space at varying levels of granularity will allow models to provide more specificity when giving low confidence predictions. It will also allow more specificity when suggesting additional information is needed. Requesting additional information and retraining is something outside of the remit of passive methods and is something that active approaches look to do, these will be discussed in the following sections.

\section{Active Approaches}\label{active}

In this category of approaches, rather than the system being self-contained and complete, it has the option to adapt as the input landscape evolves. That is, to function optimally an adaptive learning model will reconfigure itself when encountering uncertainty. Here, three splits are identified, each defining a different magnitude of reconfiguration. In active learning the reconfiguration occurs in an input-wise fashion, whereby individual data points are selected to inform model updates. In the graceful degradation scenario this might be after the model has been deployed and operate in a manner akin to lifelong or continual learning, hence both will be discussed in this area. These approaches might be applied when encountering a new object within a known domain. The second area is that of domain adaptation where the model is trained on a distribution of data and then asked to operate on a different distribution of data. This might be relevant when a models task does not change but its environment does, like a disaster response drone moving to a new landscape. Hence the model is retrained to fulfil its purpose in the new environment. Finally, in meta-learning the feature space of the model will change and the model is encountering objects that are different to that on which it was trained. Hence, retraining is necessary if the model is to be of any use. Such techniques might be applicable when a smart city system is applied to a new city or area and is required to help assist in new/different events.

\subsection{Active and Continual Learning}\label{active-continual}

In this category of approaches the models will be updated using smaller reconfigurations, such updates might be needed when seeing a particular data point which causes uncertainty in the model, like a peculiar situation for a smart city system or a particular object in a disaster response drone. Here the operator can label a particular datapoint for the model to update itself. This might update the systems knowledge of the distribution as in the active learning case, or it might begin to give the system knowledge of a new task as in continual/lifelong learning. The important distinction here is that it happens with a small number of data points, as opposed to entire distributions.

In an ideal scenario unconfident samples would be continuously labelled by the operator in a continuous manner, as in continual learning. In this field the system is faced with a continuous stream of data to learn from. A popular paradigm in this field is the online setting, where the data is passed through once and the system is required to incrementally build its knowledge of the problem. The approaches to continual learning often focus on alleviating the catastrophic forgetting problem and can be split into three categories: Regularisation-based, memory-based, and model-based. Regularisation-based approaches impose constraints on the weight updates of the model (\cite{kj2020meta}, \cite{jung2020adaptive}, \cite{mirzadeh2020understanding}), memory-based methods centre the approach on \textit{memorable} examples in the data space (\cite{van2020brain}, \cite{cong2020gan}, \cite{pan2020continual}, \cite{ke2020continual}), and model-based approaches change the architecture of the model to handle new information (\cite{singh2020calibrating}, \cite{mallya2018packnet}). A recent paper by \cite{caccia2020online} presents a novel approach that focusses on solving unseen out-of-distribution tasks. A paper by \cite{vandeven2019scenarios} presents an interesting breakdown of the applications of a continual learning system, whereby the system either has to classify an item if given a task, classify an item according to a distribution tasks, or classify both the item and the task it belongs to. 

In a deployed system all three of these scenarios could be consistent with an active gracefully degrading system. In this scenario continual learning could be implemented to gradually build a models understanding of the task spaces and data distributions within them, by continuously providing labels for low confidence classifications due to out-of-distribution errors. However, this might not always be feasible and is counterintuitive as the system is in place to automate the classification process, so it may be beneficial to also review the field of active learning. 

In modern machine learning problems annotated data is often a luxury that takes a great deal of time to produce, therefore obtaining usefully annotated data quickly will be important. In active learning, the goal of the algorithm is often to choose data that will most appropriately separate the classes and suggests that it be annotated. The two most popular paradigms in the active learning field are pool-based sampling and stream based sampling. 

In stream-based sampling, inputs are continuously given to the model and it is is asked whether to request for a label from the \textit{oracle} or label the point itself, which bears some similarities to continual learning. Some recent approaches in this area include a paper by \cite{Shah2020} where an option to reject classification is included and a paper by \cite{murugesan2017active} where an ensemble-esque approach is taken. Other recent work includes that by \cite{fujii2016budgeted} and \cite{hong2020active}.

Pool-based sampling presents a pool of unlabelled images to the model and supplies a corresponding annotation budget with which to optimally consult the oracle for annotations. It first trains on the samples allocated to it randomly, the model then classifies the remainder of the pool. The samples with the lowest confidence are then selected for annotation by the oracle, moving them to the training set. The process is repeated until test performance on a test set is sufficiently high. There are two methods of selecting samples for labelling: Singleton labelling where single points are chosen for annotation (\cite{yang2018variance}, \cite{beluch2018power}) and batch labelling where a batch of points are selected for annotation (\cite{ash2019deep}, \cite{yoo2019learning}, \cite{shi2019integrating}). Recent papers by \cite{sinha2019variational} and \cite{tran2019bayesianActive} take variational and bayesian approaches respectively, methods also discussed in \cref{epistemic-uncertainties}. Selecting samples individually can often be wasteful of resources, hence more recent progression has concentrated on the batch-based approach.

In deployed machine learning systems the oracle will refer to an operator and the system may desire manual annotation when it encounters an object it has difficulty classifying confidently. This might refer to the first vignette, whereby the model asks the operator to label an object that falls outside of the trained distribution, thereby allowing to the model to informatively update its understanding of that area of the distribution. 

\begin{figure}[h!]
\centering
\includegraphics[width=0.85\textwidth]{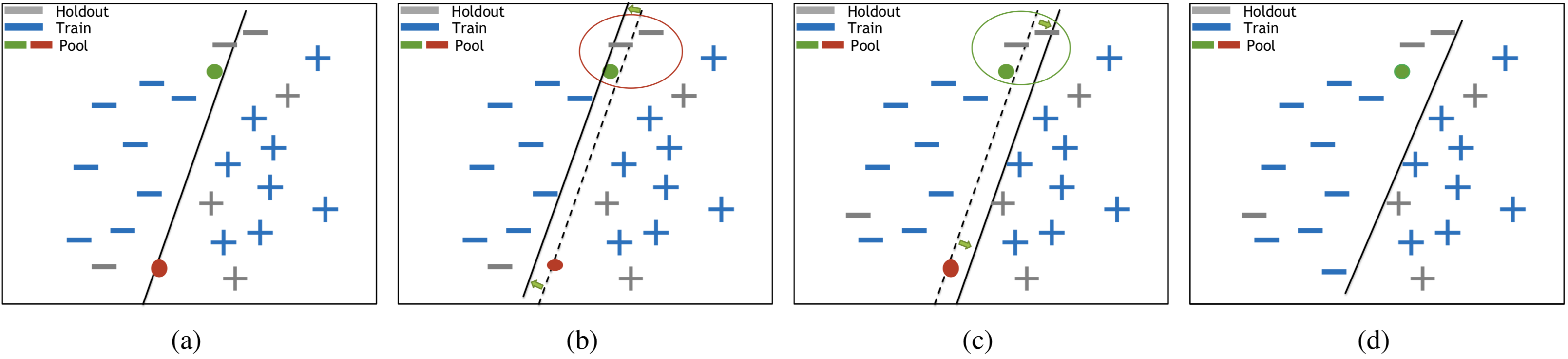}
\caption{A figure taken from the paper by \cite{aljundi2020identifying} presenting the problem of maximising confidence alone in the active learning problem.}
\label{incorrectly-labelled}
\end{figure}

In active learning the most important measure in this problem is how samples are chosen to be annotated, a common metric for this is confidence, as discussed in \cref{uncertainties} there are a variety of ways to do this. However, if a sample lies close to a class boundary and is miss-classified, minimising uncertainty in that prediction will inject error into the model. Since in the active learning scenario the model is not to know it is incorrectly classified, only that it is not confident in its prediction. Now increasing the confidence in that same prediction by picking a different label for annotation will improve the active learning loss function, but compound the underlying error on the uncertain label. This concept is explored in a recent paper by \cite{aljundi2020identifying}. A figure from the paper illustrating this issue is shown in \cref{incorrectly-labelled}.

The problem illustrated in \cref{incorrectly-labelled} is that updating according to the red sample by choosing to query the green sample will increase the generalisation error on the held-out samples, whereas querying the red sample will improve the error on the held out samples. This is the underlying methodology of the approach: The model predictions are hypothesised to be correct and the model updates are temporarily made. The effect on the generalisation error is then noted using a small held out set of annotated data-points. This allows the model to make a more informed guess as to which is incorrectly labelled, as the hypothesised model will increase generalisation error on the held out set. To minimise the computational implications of the approach a first order approximation of the method is also presented.

To evaluate the approach and its approximation, image classification and image segmentation datasets were selected. In image classification two settings are considered, one where all classes are balanced and another more realistic setting where the classes are unbalanced. Tests were performed on MNIST, K-MNIST (\cite{clanuwat2018deep}), Street View House Numbers (SVHN) (\cite{netzer2011reading}), and CIFAR-10. In the balanced setting results were close with all methods proving competitive, however the proposed approach was more efficient in computation. The method, as well as its approximation, consistently outperformed competing methods in the imbalanced setting. This was true on all but the CIFAR-10 set, where all methods generally showed the same performance. It is hypothesised in the work that this is due to low intra-class similarity in the CIFAR-10 set, thus reducing the discriminatory impact gained from individual samples. 

Such approaches could be applied to the graceful degradation problem where manual interventions from to the operator on low confidence predictions is expensive. However, when these actions are not expensive and manual annotation can be afforded more often, continual learning strategies might be preferred whereby the model learns representations as the task and input space evolves. A challenge in the field of continual learning is the catastrophic forgetting problem, where information obtained by the model is lost as it sequentially updates itself. Hypernetworks, discussed earlier in \cref{epistemic-uncertainties} to provide efficient variational inference, are also implemented in a paper by \cite{von2019continual}, now applying them in continual learning to memorise task specific weight realisations.

\begin{figure}[h!]
\centering
\includegraphics[width=0.85\textwidth]{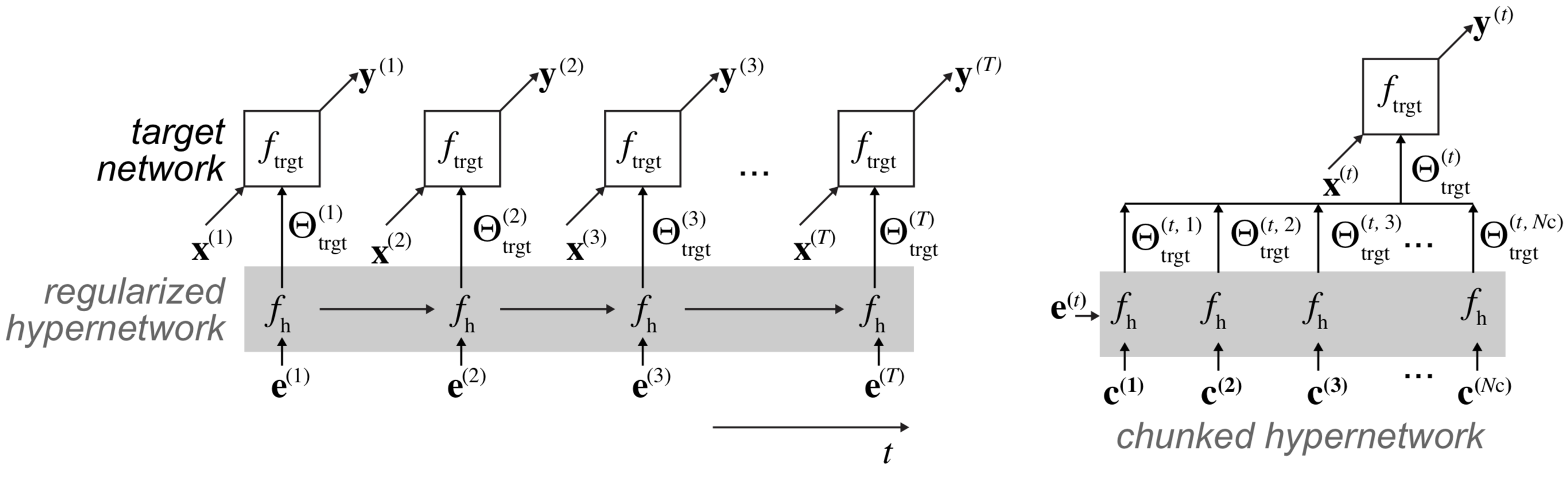}
\caption{A diagram taken from the paper by \cite{von2019continual} presenting the design of the proposed approach. The full solution shown on the left, and the chunked on the right.}
\label{continual-hypernetwork}
\end{figure}

Rather than attempting to retain the output $x$ of a model for learnt tasks: $f(x,\theta)$, the authors aim to instead retain the outputs of a \textit{metamodel}, $f_h(e,\theta_h)$, where a task embedding, $e$ is mapped to its corresponding weights, $\theta$. This marks a change in perspective on the continual learning problem, where the authors report a model which can maintain high performance on tasks without forgetting information. Hypernetworks are used here to parameterise target models and avoid the problem of catastrophic forgetting by representing these target models in the weight space using a single point, as opposed to memory intensive approaches that iterate over all previous data. To reduce the computational cost of the method the hypernetwork is implemented in an iterative manner in a method known as chunking. In this scenario the hypernetwork takes the task embedding and the particular chunk embedding as an input to produce its weight output. This is presented in \cref{continual-hypernetwork} where the hypernetwork produces models that create the desired weights, $\Theta_{trgt}$, for the input, $e$.

When evaluating the model the authors consider the same three continual learning scenarios presented in the paper by \cite{vandeven2019scenarios}. Two datasets are used to do this, MNIST and CIFAR-10/100. A common MNIST benchmark in continual learning is the permuted MNIST, whereby the model is first trained on the MNIST digits and subsequent tasks are created by applying random perturbations to the input pixels. This is designed to study the memory capacity of a continual learner and most appropriately fits the first continual learning scenario. The second benchmark using MNIST is the \textit{split MNIST benchmark} which introduces class overlap, here the digits are sequentially paired to form five binary classification tasks. This test fits the second and third continual learning scenarios, where the model is not told which task it is performing. Finally the model is evaluated on the most challenging continual learning problem, the split CIFAR-10/100 task, where the model first learns the CIFAR-10 dataset and is then presented with sets of ten classes from the CIFAR-100 dataset, which it is asked to learn in a sequential manner. The model produced state-of-the-art results on all of these continual learning benchmarks. 

It is clear such continual learning scenarios could be pertinent in the graceful degradation problem, where the system evolves with its environment. However, such approaches would require the annotation of new data in these environments. Perhaps the coalescence of active and continual learning could optimally inform which data should be annotated as this occurs. In the graceful degradation scenario this can account for unknown unknowns, provided there is intervention from the operator, as in the continual learning scenario new tasks and classes can be introduced as the learning continues. Hence, such models could be incorporated into scenarios where epistemic uncertainties are expected, if there are resources available to facilitate the learning in the system. 

Sometimes such learning will be too slow to undertake after the deployment of a system and such environmental changes can be anticipated. In this scenario it may be beneficial to have a system in place that allows the quick adaptation of a system when its environment changes. This is where techniques such as domain adaptation and meta-learning could play an important role in a gracefully degrading system. Hence, these will be discussed in the next sections. 


\subsection{Domain Adaptation}\label{domain-adaptation}

In domain adaptation (DA) the goal is to create a system that is trained on one distribution, but operates in the context of another distribution. There exist two types of DA at the broadest level: Single-step and multi-step. In single step domain adaptation the model switches domain through a single continuous adaptation. In multi-step domain adaptation the model switches domains via an intermediate domain, once finding an appropriate intermediate domain it is easy to see the method becomes a single-step problem. Once in the single-step domain, the methods can be split again into homogenous and heterogenous. The most notable difference between these are the feature spaces in the training and target distributions. In the homogenous regime the feature spaces are the same and in heterogenous the feature spaces differ, arguably making it more akin to transfer learning. In homogenous domain adaptation the only difference is in the distributions of the sets of data, as the source and target domain share a relationship between one another. Hence, this is closely related to the graceful degradation problem, as like in the second vignette domain adaptation is about the model understanding data is outside of its training domain and re-training for the new domain. This might be a change in environment for a disaster response drone, or change of season in a smart city system.

Homogenous domain adaptation can again be split into: Discrepancy-based approaches (\cite{balaji2019normalized}, \cite{chen2020HoMM}, \cite{Zellinger2019}, \cite{Li20DCAN}), adversarial-based (\cite{Long2018}, \cite{cui2020gradually}, \cite{Haoran_2020_ECCV}, \cite{jiang2020implicit}, \cite{dada}, \cite{jiang2020implicit}), and reconstruction-based (\cite{gong2019dlow}, \cite{bousmalis2016domain}, \cite{srdc}). In discrepancy-based approaches the deep network is fine-tuned to diminish a discrepancy caused by the domain shift. This discrepancy can be based on a variety of criteria, such as class labels, statistical distributions, model architecture, and geometric signatures. In adversarial-based approaches domain discriminators encourage domain confusion through an adversarial objective, this might use generative or non-generative approaches. Reconstruction-based approaches use data reconstruction as an auxiliary task ensuring feature invariance. Of course, there are some approaches that do not fit into this segregation, such as the heuristic approach shown in the paper by \cite{cui2020hda} and a semi-supervised approach presented by \cite{zhang2020label}. It is also worth noting these approaches share some resemblance to transfer learning and many techniques will be seen in both fields, the main difference being in domain adaptation there often do not exist labelled examples in the target domain. 

Many of these approaches assume there exists labelled examples in the new domain, indeed all of the approaches outlined above can be undertaken in a supervised, semi-supervised, or unsupervised manner. Clearly having labelled examples in the source domain is a luxury in this problem and having unlabelled samples will be more common, hence recent papers utilising unlabelled data will be highlighted in this section.

\begin{figure}[h!]
\centering
\includegraphics[width=0.7\textwidth]{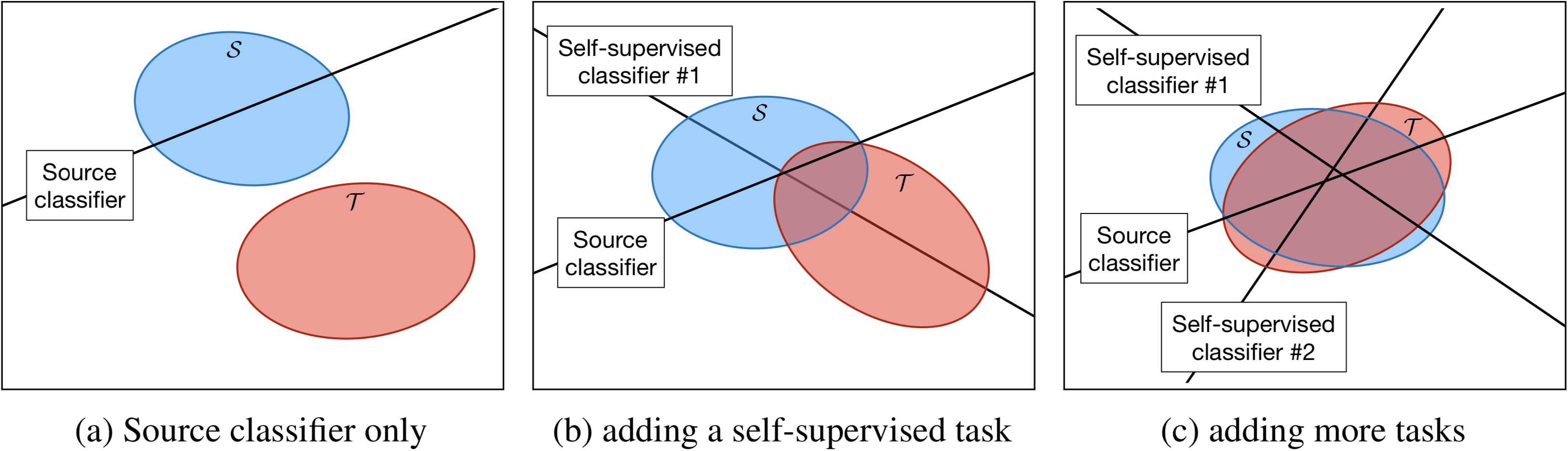}
\caption{A figure from a paper by \cite{sun2019unsupervised} demonstrating how self-supervision tasks help align domains in the representation space.}
\label{SSDA}
\end{figure}

\cite{sun2019unsupervised} present an unsupervised method of domain adaptation that utilises self-supervised learning to enable transfer between domains. Presented, is a model that learns supervised representations for a classification task, whilst performing a self-supervised task on both the source and target domain. The main contribution of the paper is presenting self-supervision tasks that are suitable for domain adaptation. Three tasks are identified: Rotation prediction, flip prediction, and patch localisation. In rotation prediction the model is trained to recognise to what angle an image has been rotated. In flip prediction whether the image has been flipped or not. Finally, in patch localisation the model is trained to return the correct coordinates on a 2x2 square for each of four quadrants of a segmented image. They also argue that multiple self-supervised tasks will help merge the target and source domains more effectively, this is demonstrated graphically in a figure from the paper shown in \cref{SSDA}.

To train the system, $K$ self supervised tasks were selected creating $K+1$ loss functions including the supervised prediction task on seen data. This creates $K+1$ classification heads ($h_{0,1...k-1,k}$) and a common feature extractor, the parameters of which are learned by the system. In this case the feature extractor is a deep CNN and the classification heads are simple linear layers. At test time all heads except $h_0$ are discarded, as $h_0$ is the classification head. 

The model was tested on seven object recognition benchmarks, on four of which state-of-the-art accuracies was achieved when utilising all three self-supervision techniques. They also test their method on object segmentation but find the self-supervised tasks are designed for object detection, thus the method is not as effective. However, the paper presents state-of-the-art domain adaptation when there exists no labels in the target domain, but there is unlabelled data. It is also suggested that this technique may be effective where the amount of data in the target domain is small. 

\begin{figure}[h!]
\centering
\includegraphics[width=0.65\textwidth]{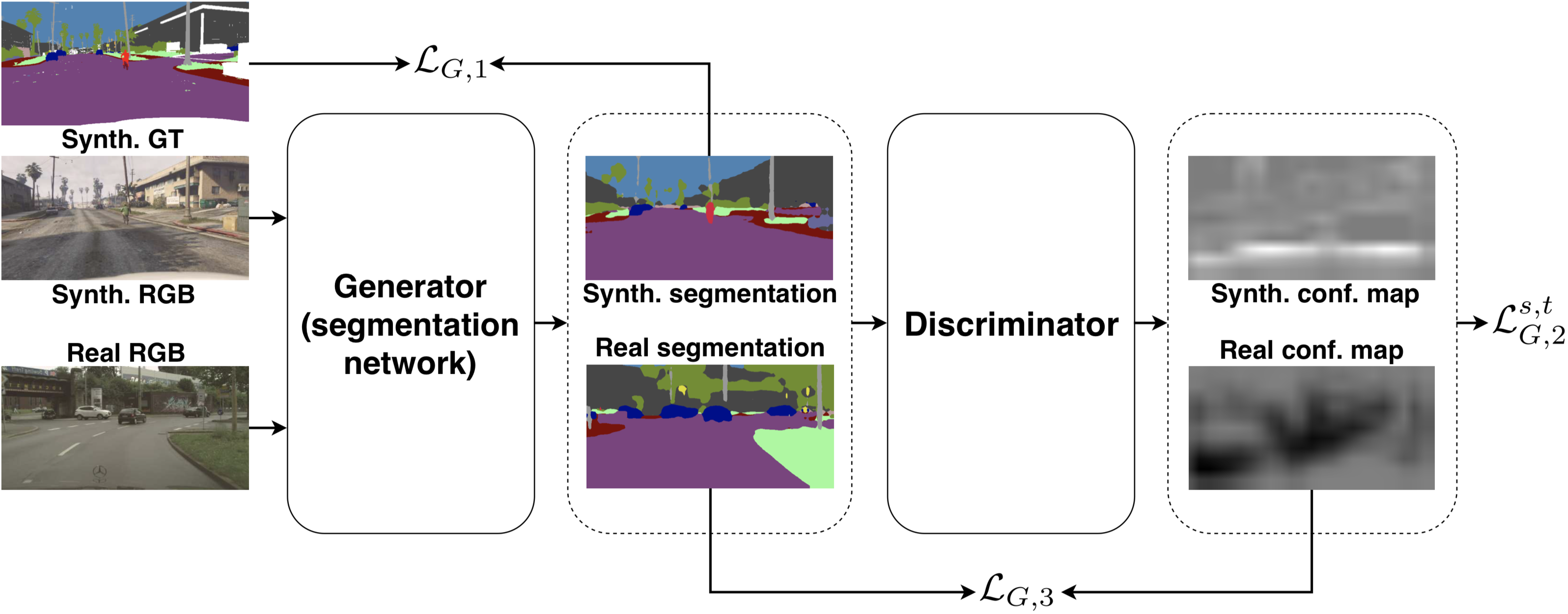}
\caption{A figure from a paper by \cite{Biasetton_2019_CVPR_Workshops} demonstrating how the approach is structured.}
\label{unsupervised-segmentation}
\end{figure}

Another unsupervised method that performs better on image segmentation is presented by \cite{Biasetton_2019_CVPR_Workshops}. In the paper the authors propose a three component strategy, which uses supervised, adversarial, and a self-learning strategy. The supervised component of the model uses a synthetic dataset, GTA5 (\cite{richter2016playing}) and for each scene produces a class probability label in a pixel wise manner, which is trained using a cross-entropy loss function. The second part of the model is a discriminator which aims to distinguish between generated segmentations and the ground truth segmentations, this is again done in a pixel-wise manner. Since segmentations can also be generated for the target domain, this methodology utilises target domain information in an unsupervised way, as the ground truth information is only available for the synthetic dataset. The output of the discriminator is used to create a confidence map of the segmentation, allowing a self-training procedure which ensures the predictions are deemed more reliable when the discriminator marks generated samples as ground truth more frequently. The structure of their approach is shown graphically in \cref{unsupervised-segmentation}.

Two models were trained, one on the GTA5 dataset and one on SYNTHIA\footnote{SYNTHetic collection of Imagery and Annotations} (\cite{ros2016synthia}), both are synthetic datasets. The model was then evaluated on the Cityscapes dataset (\cite{cordts2016cityscapes}). On average, performance on both datasets was comparable to or exceeded that of competing approaches, if falling short on a few classes. 

Domain adaptation generally requires the existence of known unknowns, but is effective in overcoming the difficulty of passing between domains when the target domain lacks labelled data. Hence, the graceful degradation problem of moving outside of the training domain could be solved by using domain adaptation strategies on target domain information. Like in the examples above this is particularly useful when there are sufficient unlabelled instances of out-of-distribution data, that share the same feature space as the training data. It is clear these approaches would be used in the case of aleatoric predictive uncertainties. Such methods can be intensive during the training phase and will require large amounts of computational resources. Hence, whilst producing robustly trained networks quick adaptation might not be achievable in this paradigm. This is something meta-learning can facilitate.

\subsection{Meta-Learning}\label{meta-learning}

\begin{figure}[h!]
\centering
\includegraphics[width=0.85\textwidth]{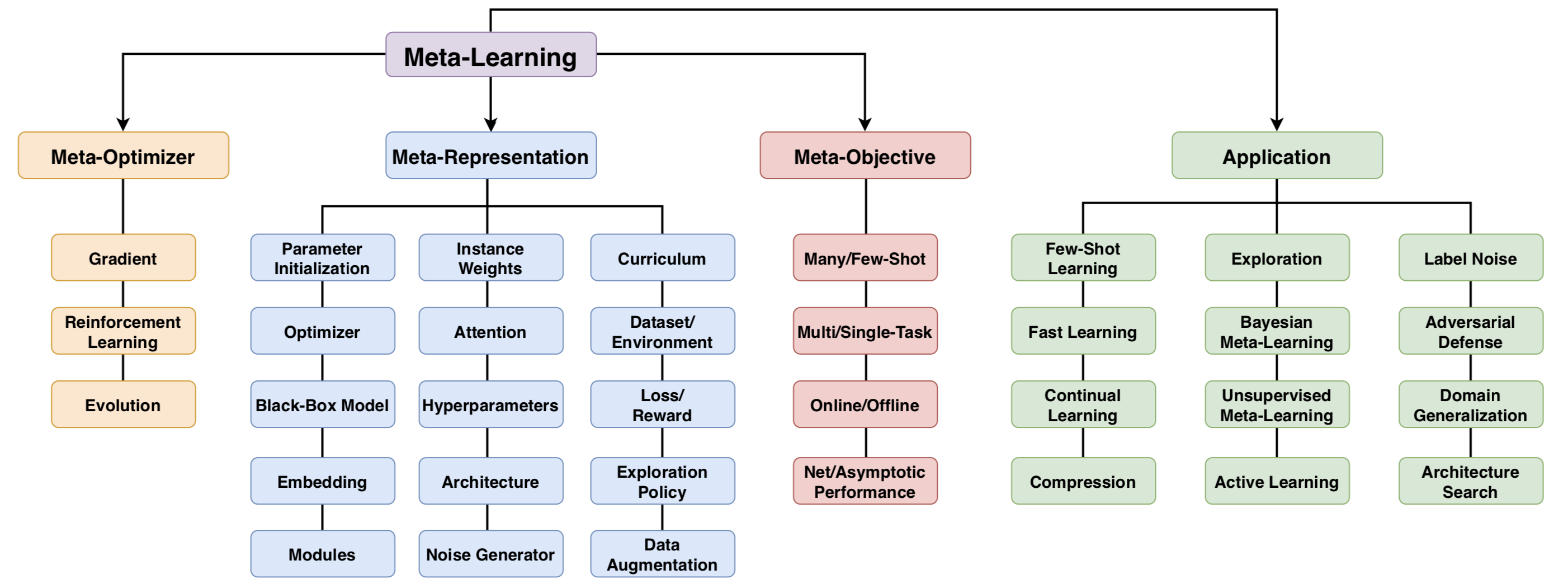}
\caption{A figure from a survey paper by \cite{hospedales2020metalearning} demonstrating a taxonomy of approaches in meta-learning.}
\label{meta-learning-taxonomy}
\end{figure}

Meta-learning is often referred to as the process of \textit{learning to learn}, this is because the models are often trained and tested on a number of tasks and datasets. Here, the key evaluative measure is the performance of a single model across a range of tasks. A recent survey paper by \cite{hospedales2020metalearning} proposes splitting the field across three axes that segregate approaches based on their meta-representations, meta-optimisers, and meta-objectives. These can be likened to \textit{what} they learn, \textit{how} they learn, and \textit{why} they learn, respectively. A taxonomy taken from the paper, as well as various applications of meta-learning, is shown in \cref{meta-learning-taxonomy}.

Perhaps the most relevant parts of this diagram in the graceful degradation problem are the meta-representation and application parts, that is how they are implemented and where they are used. Some stand out applications for the graceful degradation include fast, continual, active, and few-shot learning, domain generalisation, and adversarial defence. Most of these applications have been discussed in this survey, suggesting meta-learning is a problem that is at the heart of adapting models in the graceful degradation environment.

Meta learning in the graceful degradation problem is most applicable to the quick adaptation of models upon encountering data outside of their training distribution, this might be fast video segmentation of new object as in \cite{Thulasidasan2019a}. But, if for example a change of environment or feature space is expected, the operator may have sets of training data available to quickly retrain the model. This would be the few-shot learning environment. To that end, the state-of-the-art in this area will be discussed. This area of the field can generally be split into two approaches: Gradient-based, and memory-based. 

Memory based approaches utilise the memory of a recurrent neural network (RNN) to directly parameterise an update rule for the few shot problem (\cite{andrychowicz2016learning}, \cite{li2016learning}, \cite{chen2017learning}, \cite{ravi2016optimization}). Theoretically, RNNs are universal function approximators, hence should be able to represent any learning rule. In gradient based approaches the goal is to find an initialisation that can be adapted for a number of tasks through gradient descent. Papers by \cite{finn2017model} and \cite{flennerhag2018transferring} provided strong frameworks for this approach and have since been expanded upon in many papers (\cite{antoniou2018how}, \cite{behl2019alpha}, \cite{yao2019hierarchically}, \cite{Rajeswaran2019}) and prompted unsupervised approaches as shown in a paper by \cite{hsu2018unsupervised}. Recent work such as that by \cite{lee2018gradient} has emphasised the use of preconditioning to more directly control the gradient descent process, merging the two approaches in the field.

Memory based approaches lack a strong inductive bias on which to base an update rule, making them hard to train. While gradient-based methods seek to find a strong inductive bias on which to make updates, but restrict knowledge transfer to the initialisation of the model. Hence, there is motivation to unify both approaches in the field, a recent example of such a unification is presented in a paper by \cite{Flennerhag2020Meta-Learning}. 

\begin{figure}[h!]
\centering
\includegraphics[width=0.45\textwidth]{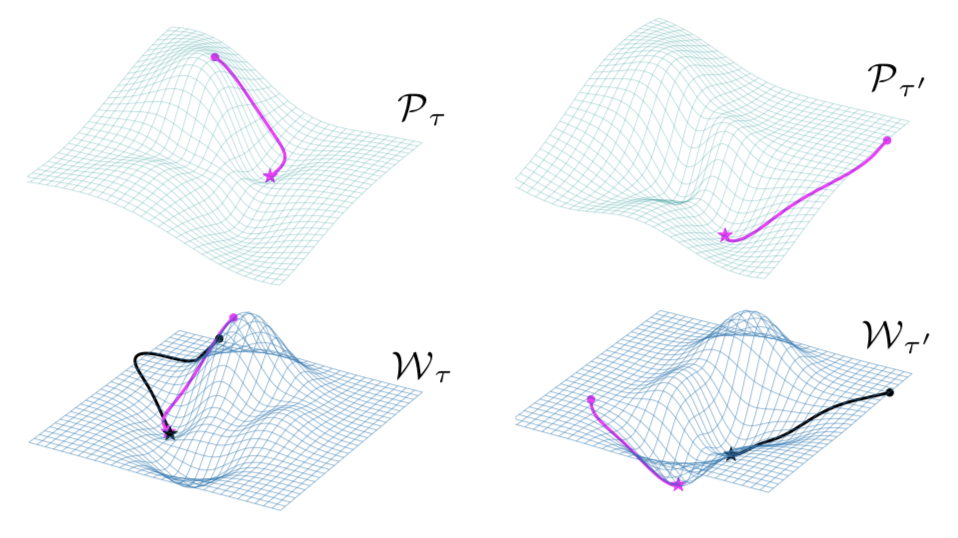}
\caption{A figure from the paper by \cite{Flennerhag2020Meta-Learning} comparing the loss landscapes of warped and non-warped representations.}
\label{warpgrad}
\end{figure}

The paper presents a method that updates the initialisation of the model as well as the way in which it makes gradient updates, through preconditioning. Here, preconditioning is parameterised by defining a subspace with layers in the network being dedicated to warping the activation space\footnote{Consider the output of each layer in the neural network as a vector, the set of vectors for each layer will form a space, known as the ‘activation space’}. This parameterises the update behaviour whilst also allowing an optimised initialisation to be found in the model. A representation of preconditioning is shown in \cref{warpgrad}, here the standard loss space, $\mathcal{W}$, is transformed to a simpler space, $\mathcal{P}$, for which it is easier to locate the minimum via gradient descent. This is demonstrated by the black and magenta update paths, the magenta path being the one informed by the warp layers. This is shown for two different tasks, $\mathcal{T}$ and $\mathcal{T}'$.

The approach is evaluated using the popular Omniglot dataset, which is a dataset containing a variety of alphabets for character recognition (\cite{lake2015human}). To meta-train, the model is trained on one of these alphabets and tested on an unseen subset of the same alphabet. The testing losses are then backpropagated to the initialisation of the model, which includes the gradient guiding warp layers. The effectiveness of the model is evaluated on a separate set of alphabets reserved for evaluation and never meta-trained on. This avoids fitting to the evaluation tasks. The evaluation procedure is performed in the same manner as training: The meta-learnt model is trained on an alphabet and then tested on a subset of the same alphabet. The difference being the model does not see the alphabet until the evaluation stage. The approach was also evaluated on a number of other datasets including mini and tiered-ImageNet (\cite{ren2018meta}), in which the method is evaluated on its ability to scale beyond few-shot classification. The work outperformed all existing approaches on all benchmarks.

Meta-learning is a field that is placed at the heart of the active approaches to the graceful degradation problem. In this setting it can be likened to the heterogenous setting of domain adaptation, where the task is changing as opposed to the domain. Whilst the paper discussed is one that uses meta-learning for few shot classification, it is clear from \cref{meta-learning-taxonomy} that the uses of meta-learning approaches extend far beyond this, be it in domain generalisation, or active learning. However, meta-learning models can be very resource intensive to instantiate, due to the necessity to train on large amounts of data. Unsupervised meta-learning can alleviate the necessity for labelled data, but cannot remove the requirement for large amounts of data. Hence, while these methods are effective in creating effective and adaptive models, they require known unknowns and large amounts of data, limiting the use cases for such approaches. The same can be said for active and passive approaches to the problem in general, understanding when one is more applicable than the other and the constraints that can govern these decisions are important considerations to be made. To this end, applications of the approaches in this survey will be discussed in the next section.


\section{Comparison and Application of Discussed Approaches} \label{applications}

The previous sections briefly discuss the state-of-the-art in the various areas of the graceful degradation problem, discussing some interesting papers in more detail. Here, the application of such approaches will be discussed and where particular constraints might prompt the use of certain techniques. First, passive and active approaches will be compared and contrasted. 

The most important factor in both passive and active approaches is uncertainty, that is, both sides of the graceful degradation problem are concerned with how they deal with uncertainty. Therefore, obtaining accurate uncertainties, both in the aleatoric and epistemic case, will be important. Where they differ is how they then manage that uncertainty. Passive approaches handle the uncertainty in a self-contained manner, naturally this limits the domains in which they can be used. This means once they are deployed updates cannot be made without to them without recalling the system. In contrast, active approaches seek to benefit from additional information and are updated according to changes in the environment or task. However, this would introduce additional difficulties when following verification and validation protocols on deployed software.

This means passive approaches will likely require less maintenance once deployed, but full functionality will likely be lost once leaving the distribution. As discussed, this can be alleviated in some way through using zero-shot learning techniques to give the model knowledge outside of the trained distribution. Furthermore, hierarchical classification models present a middle ground by offering a coarser-grained classification. However, full functionality will still be lost regardless of how epistemic uncertainty is handled. Active approaches on the other hand, whilst needing updates to ensure they are trained in the operating domains, will not lose functionality when changing environments or tasks. This means they will require ongoing maintenance as they operate across environments, but will be more adaptable than passive systems. These updates can happen at a range of magnitudes be it with individual data points, or new tasks. 

A benefit to uncertainty communicating systems and to a lesser extent, hierarchical classification systems, is that they will often be effective on unknown unknowns. This means they do not require additional information to employ their functionality on out-of-distribution data. However, in active approaches changing tasks or domains will always occur with known unknowns. Hence, these methods will be undermined by uncertainties arising due to unknown unknowns, as there is no way for the model to adapt them. Small adaptations like in active and continual learning have the potential to adapt to unknown unknowns, provided the operator is able to label the unknown datapoint correctly. 

A variety of supervision strategies are used across the breadth of the graceful degradation problem, with particular branches of the problem seeing more usage of one type than others. For example in domain adaptation and zero-shot models, unsupervised learning is used more often than fully supervised. As the existence of labelled data in the target environment will often negate the need for the approach in the first place. In contrast out-of-distribution detection models in uncertainty based methods will often use supervised approaches, as knowledge of out-of-distribution data will cause data leakage thus undermining the efficacy of the method. Hence, when considering the most appropriate supervision strategy the use case should be considered as well as the data that is available, as some approaches are better suited to certain supervision strategies. 

Each method has benefits over the other, passive approaches will be better suited when access to the system is limited and the environment in which it operates is not likely to change. Active approaches will be better suited in scenarios where the system is likely to encounter a number of environments, which might be known to an operator beforehand. Ultimately, the usage of passive and active approaches will be determined by the computational resources available to a system, the data available to train it, and its use case. To that end, some likely scenarios and limiting cases will be considered for the remainder of the section.

\paragraph{Unknown unknowns} 

When function on unknown unknowns is needed, this will limit some of the gracefully degrading implementations available. Passive approaches that express uncertainty and obtain accurate estimations of them will be useful, some hierarchical classification systems might be available too, provided the target object is not too far out-of-distribution. In terms of active approaches both meta-learning and domain adaptation require preparation for the target distributions, hence limiting the uses to problems with known unknowns. Active and continual learning on the other hand could be used on unknown unknowns, provided the operator is able to label the unknown object. For example, in stream-based active learning the system is faced with new data continuously, some of this it will be able to classify, but some may need labelling by the operator. The same could be said for continual/lifelong learning. Therefore, the approaches that could be used when you anticipate unknown unknowns are: Active learning, continual learning, hierarchical classification, and out-of-distribution detection/accurate epistemic uncertainty models.

\paragraph{Low data in the target domain} 

When there exists a small amount of data in the target domain some approaches might be ruled out. For example some uncertainty based methods will lose functionality if they cannot be sufficiently trained in the new domain and simply return uncertainty. Zero-shot and hierarchical classification methods might be useful, but it would be more appropriate to use methods that more efficiently use the labelled data. Therefore, active approaches might be preferred in this case. For example, meta-learning could be used in the few-shot environment to quickly adapt a model to classify objects in the target domain. Domain adaptation models could be used, but the small amounts of data would likely prevent them from achieving the desired performance in the target domain, the same might be said for active and continual learning. However, if the change between domains is small active approaches may be useful as they can build upon existing representations of the data space. Therefore, when limited in the amount of data in the target domain, active approaches will likely be preferred over passive, namely active learning and meta-learning. It is also important to consider the existing knowledge of the system when selecting between these approaches.

\paragraph{Data is not in the target domain}

In this scenario the desired functionality of the model should be considered. For example, if the aim is to classify objects in the new target domain, then zero-shot classification will be preferred as it will provide the model with a baseline understanding of the new area. If the aim is to express uncertainty when encountering the new domain, uncertainty communicating models may still be useful. Furthermore, hierarchical classification models could be useful in this scenario, providing coarse level labels on data in the new domain. However, active approaches would be rendered ineffective until data could be collected in the new domain, even then, this data would be unlabelled and will require annotation by an operator. Therefore, the most appropriate methods when there is no data in the target domain are: Zero-shot classification, hierarchical classification, and out-of-distribution detection/accurate epistemic uncertainty models.

\paragraph{There only exists unlabelled data}

In the scenario that there is only unlabelled data in the new domain active approaches will be more useful than passive approaches. For example, domain adaptation could efficiently leverage the new data, provided the labels are the same across both domains. Meta-learning also has unsupervised methodologies, hence these could also be used to adapt the model. If an operator is available to annotate a selection of data, pool-based active learning could be implemented. Some uncertainty aware approaches that use unlabelled data could be employed. For example, unsupervised out-of-distribution techniques could be used so the system is aware that it has left the intended distribution. But such functionalities can be achieved without leveraging additional data. Hence, to make the most use of the data available an active approach would likely be preferred. But, the most appropriate active approach will be determined by the particular scenario and the availability of resources.



\section{Open Challenges and Conclusion} \label{conclusion}

Throughout this survey the literature of various fields related to graceful degradation have been discussed. It has shown that graceful degradation can be built into cutting edge AI enabled systems, by implementing some of the techniques discussed. However, that is not to say these fields are all optimally suited for graceful degradation. Hence, there are various open areas in which further development could be made to better enable gracefully degrading machine learning systems. In this section a number of these will be identified and conclusions of the survey will be presented.

Bayesian inference methods are very successful in giving suitable predictive uncertainty with its predictions, however they have high inference costs relative to fully deterministic models. Dynamic inference methods have shown to have impressive results in making complex models more efficient (\cite{huang2017multi}, \cite{wang2017skipnet}, \cite{teerapittayanon2016branchynet}, \cite{9028245}). Unlike other power saving methods, dynamic inference allows the full model to be used when necessary, meaning predictive uncertainty could be preserved. Efficiency at the inference stage of bayesian models is something that has not seen a lot of development, dynamic inference strategies could provide this. 

In hierarchical classification hyperbolic embedding spaces have recently been implemented to better represent the hierarchy of classes. Such embedding spaces could be implemented in other models where the preservation of distances are important in the embedding space. Furthermore, as shown in a paper by \cite{passalis2020efficient}, work has recently been done with branched neural networks to facilitate hierarchically minded representations. 

Branched neural networks are an implicit and powerful form of dynamic inference that allow predictions of improving predictive uncertainty at the expense of increasing computational expense. This allows for an input adaptive system that can quickly and confidently classify easy samples, but will use more computational resources on hard samples. Hierarchical bayesian inference would compliment such models, allowing for confident and efficient inference at varying levels of class hierarchy, or alternatively, fast confident bayesian inference utilising early exiting strategies. 

As discussed, uncertainty is the core property of a gracefully degrading system, many of the methods in this survey fail to include uncertainty measures in their analysis. For example, fields such as zero-shot learning, meta-learning, and domain adaptation concentrate predominantly on the accuracy of predictions, as opposed to appropriate predictive confidence. Whilst accuracy is an important metric for such areas, the confidence of incorrect predictions should also be considered and should be minimised where possible. Hence, the uncertainty aware methods discussed in \cref{epistemic-uncertainties} should be transferred to other areas if they are to be moved to the deployable domains.

There exist a lot of similarities between domain adaptation and GZSL, as the main goal is for the system to work on seen and unseen classes. The only difference between the two settings is that domain adaptation gets access to unlabelled examples of the classes. There does exist a field in the zero-shot learning field, \textit{transductive} zero-shot learning, where the model has access to unlabelled examples of the unseen classes. It follows that this field is identical to domain adaptation, therefore, there is opportunity for the two areas to take inspiration from one another. 

Stream-based sampling his perhaps the most relevant active learning approach as it most appropriately reflects the operator--system scenario, whereby the system will consult an oracle when it lacks confidence. However, this area is not particularly active, being replaced by pool-based sampling. If graceful degradation is to progress further, active learning approaches should be evaluated in both the pool-based and stream-based sampling settings. Hence, some future work would be a comprehensive evaluation of all state-of-the-art active learning algorithms, across both settings, to understand which techniques are the most applicable to given scenarios. 

This survey began by understanding what is meant by graceful degradation and formally defined the graceful degradation problem in machine learning. It was defined as the notion of minimising the effects of out-of-distribution errors. These effects often take the form of increased errors and more importantly, over-confidence in these erroneous predictions. Following this some vignettes were considered to better understand the applications of gracefully degrading vision systems and where they might be useful.  

The various approaches to the graceful degradation problem were then arranged into a taxonomy to better understand the various approaches in the field and how they can be arranged. It was understood that appropriate uncertainty is a core idea of the graceful degradation problem. Hence, uncertainty was explored as well as various uncertainty and confidence metrics. A novel fundamental split was identified in the approaches Passive and active. This allowed the graceful degradation problem to be analysed in an application-centric manner.

Using the taxonomy a comprehensive review of the field was undertaken, first starting with the passive approaches, including methods of obtaining accurate uncertainties, zero-shot classification, and hierarchical classification models. Then active approaches were considered, including active and continual learning, meta-learning, and domain adaptation. Finally some future areas of research were considered, namely suggesting that branched neural networks could interact with the graceful degradation problem in an effective manner. 

This paper has formalised the graceful degradation problem and surveyed the relevant areas of the machine learning landscape. If machine learning is ever to be deployed in safety critical systems it is clear that the inclusion of graceful degradation in the design methodology is imperative. This survey has aimed to communicate the importance of the problem and prompt the development of machine learning strategies that are aware of graceful degradation.

\section{Acknowledgements}
	This work was supported in part by the following: 
\begin{itemize}
\item My supervisors Professor Steve Gunn and Dr Sebastian Stein for providing helpful discussions on a weekly basis
\item The Alan Turing Institute for providing feedback on the development of the survey on a regular basis
\item The Defence Science and Technology Laboratory for supporting the project and providing feedback on the report
\item The EPSRC Centre for Doctoral Training in Machine Intelligence for Nano-electronic Devices and Systems, funded by the UK Engineering and Physical Sciences Research Council and the University of Southampton 
\end{itemize}

\bibliography{Papers}
\bibliographystyle{iclr2020_conference}
\newpage
\begin{appendix}

\section{Github Repositories}

Here, some of the papers with accompanying GitHub repositories are detailed with accompanying repositories.

\subsection{Passive Approaches}
\subsubsection{Accurate Epistemic Uncertainties}

Data/Training-Based:
\begin{description}
\item[Uncertainty-aware deep classifiers using generative models:]{\url{https://muratsensoy.github.io/gen.html}}
\item[Improving confidence estimates on unfamiliar examples:]{\url{https://github.com/lizhitwo/ConfidenceEstimates}}
\end{description}

Model-Based:
\begin{description}
\item[Implicit weight uncertainty in neural networks]{\url{https://github.com/pawni/BayesByHypernet/}}
\item[Weight uncertainty in neural networks:]{\url{https://github.com/saxena-mayur/Weight-Uncertainty-in-Neural-Networks}}
\item[Bayesian Layers:]{\url{https://github.com/google/edward2}}
\item[Bias-reduced uncertainty estimation for deep neural classifiers]{\url{https://github.com/geifmany/uncertainty_ICLR}}
\end{description}

Post-Hoc:
\begin{description}
\item[Beyond temperature scaling- Obtaining well-calibrated multi-class probabilities with Dirichlet calibration]{\url{https://dirichletcal.github.io/}}
\item[Liklihood ratios for out of distribution detection:]{\url{https://github.com/google-research/google-research/tree/master/genomics_ood}}
\item[Posterior Network: Uncertainty Estimation without OOD Samples via Density-Based Pseudo-Counts]{\url{https://github.com/sharpenb/Posterior-Network}}
\end{description}

\subsubsection{Zero-shot Classification}

\begin{description}
\item[Creativity inspired zero-shot learning]{\url{https://github.com/mhelhoseiny/CIZSL}}
\item[Hyperbolic Visual Embedding Learning for Zero-Shot Recognition:]{\url{https://github.com/ShaoTengLiu/Hyperbolic_ZSL}}
\end{description}

\subsection{Hierarchical Classification}
\begin{description}
\item[A Fully Hyperbolic Neural Model for Hierarchical Multi-Class Classification]{\url{
https://github.com/nlpAThits/hyfi}}
\item[Hyperbolic graph convolutional neural networks:]{\url{ https://github.com/facebookresearch/ hgnn}}
\item[Making Better Mistakes- Leveraging Class Hierarchies with Deep Networks:]{\url{https://github.com/fiveai/making-better-mistakes}}
\item[Nested Learning for Multi-Granular Tasks]{\url{https://github.com/nestedlearning2019}}
\end{description}

\subsection{Active Approaches}

\subsubsection{Active and Continual Learning}

Continual Learning:
\begin{description}
\item[Meta-Consolidation for Continual Learning]{\url{https://github.com/JosephKJ/merlin}}
\item[Brain-inspired replay for continual learning with artificial neural networks:]{\url{https://github.com/GMvandeVen/brain-inspired-replay}}
\item[Memory Replay GANs- learning to generate images from new categories without forgetting:]{\url{https://github.com/WuChenshen/MeRGAN}}
\item[Continual Learning of a Mixed Sequence of Similar and Dissimilar Tasks]{\url{https://github.com/ZixuanKe/CAT}}
\item[Continual learning with hypernetworks]{\url{https://github.com/chrhenning/hypercl}}
\end{description}

Active Learning:
\begin{description}
\item[Bayesian Generative Active Deep Learning github]{\url{https://github.com/toantm/BGADL}}
\item[Variational Adversarial Active Learning]{\url{https://github.com/sinhasam/vaal}}
\end{description}

\subsection{Homogenous Domain Adaptation}

Discrepancy based:
\begin{description}
\item[DLOW- Domain Flow for Adaptation and Generalization:]{\url{https://github.com/ETHRuiGong/DLOW}}
\item[HoMM: Higher-order Moment Matching for Unsupervised Domain Adaptation]{\url{https://github.com/chenchao666/HoMM-Master}}
\end{description}

Adversarial-Based:
\begin{description}
\item[Classes Matter- A Fine-grained Adversarial Approach to Cross-domain Semantic Segmentation:]{\url{https://github.com/JDAI-CV/FADA}}
\item[Implicit Class-Conditioned Domain Alignment for Unsupervised Domain Adaptation]{\url{https://github.com/xiangdal/implicit_alignment}}
\end{description}

Reconstruction-Based:
\begin{description}
\item[Gradually Vanishing Bridge for Adversarial Domain Adaptation]{\url{https://github.com/cuishuhao/GVB}}
\item[Unsupervised Domain Adaptation via Structured Prediction Based Selective Pseudo-Labeling]{\url{https://github.com/hellowangqian/domain-adaptation-capls}}
\end{description}

\subsection{Meta-learning for few-shot}

Memory-Based:
\begin{description}
\item[Optimization as a Model for Few-Shot Learning]{\url{https://github.com/twitter/meta-learning-lstm}}
\end{description}

Gradient-Based:
\begin{description}
\item[Model-Agnostic Meta-Learning for Fast Adaptation of Deep Networks]{\url{github.com/cbfinn/maml}}
\item[Meta-Learning with Warped Gradient Descent]{\url{https://github.com/flennerhag/warpgrad}}
\item[Meta-Learning with Implicit Gradients]{\url{http://sites.google.com/view/imaml}}
\item[Gradient-Based Meta-Learning with Learned Layerwise Metric and Subspace]{\url{https://github.com/yoonholee/MT-net}}
\end{description}
\end{appendix}

\end{document}